%% file: main.tex
\documentclass[sigplan,screen]{acmart}
\usepackage{booktabs} % For formal tables
\usepackage[english]{babel}
\usepackage{xcolor}
\usepackage{xspace}
\usepackage{colortbl}
\usepackage{pifont}
\usepackage{subfigure}
\usepackage{graphicx}
\usepackage{booktabs}
\usepackage{threeparttable}
\usepackage{times}
\usepackage[textfont={it}, font={small,bf}, skip=1pt]{caption}

\usepackage{multirow}
\usepackage{fancybox} % for Sbox
\usepackage{amsmath}
\usepackage{hhline}
\usepackage{amsthm}
\usepackage{comment}
\usepackage{tabularx}
\usepackage{bbm}
\usepackage{amsfonts}
\usepackage{pythonhighlight}
\usepackage[inline,shortlabels]{enumitem}
\usepackage{makecell}
\usepackage{algorithmicx}
\usepackage[linesnumbered,ruled,noend,scleft]{algorithm2e}
\usepackage{float}
\usepackage{stfloats} 
\usepackage{xcolor}

\definecolor{codegray}{rgb}{0.5,0.5,0.5}
\definecolor{codegreen}{rgb}{0,0.6,0}
\definecolor{codepurple}{rgb}{0.58,0,0.82}
\definecolor{backcolour}{rgb}{0.95,0.95,0.92}

\acmYear{2025}\copyrightyear{2025}
\setcopyright{cc}
\setcctype[4.0]{by}
\acmConference[SOSP '25]{ACM SIGOPS 31st Symposium on Operating Systems Principles}{October 13--16, 2025}{Seoul, Republic of Korea}
\acmBooktitle{ACM SIGOPS 31st Symposium on Operating Systems Principles (SOSP '25), October 13--16, 2025, Seoul, Republic of Korea}
\acmDOI{10.1145/3731569.3764829}
\acmISBN{979-8-4007-1870-0/25/10}

\def\ie{{i.e.}}
\def\eg{{e.g.}}

\def\inline1x{Model-S}

\def\name{IC-Cache\xspace}
\usepackage{tikz}

\newcommand*\blackcircled[1]{\tikz[baseline=(char.base)]{
            \node[shape=circle,fill,inner sep=1pt] (char) {\textcolor{white}{#1}};}}

\newcommand{\update}[1]{\textcolor{black}{#1}}

\usepackage{tcolorbox}
\tcbuselibrary{skins, breakable} % Needed for styling and page breaks
\usepackage[T1]{fontenc}

\newtcolorbox{llmprompt}[2][]{%
    enhanced,                   % Use enhanced skin features
    colback=blue!5!white,       % Background color: very light blue
    colframe=blue!75!black,     % Frame color: dark blue
    fonttitle=\bfseries\scshape,% Title font: bold small caps
    fontupper=\ttfamily,        % Set the main content font to monospace (Bera Mono)
    title={#2},                 % Set the title using the mandatory argument
    attach boxed title to top left={yshift=-0.1in, xshift=0.15in}, % Position title
    boxed title style={
        colback=blue!75!black, % Title box background: dark blue
        colframe=blue!75!black % Title box frame: dark blue
    },
    before upper={\vskip2mm}, % Add some space below the title box,
    #1
}

\newtcolorbox{llmexamples}[2][]{%
    enhanced,                   % Use enhanced skin features
    colback=yellow!5!white,       % Background color: very light blue
    colframe=yellow!75!black,     % Frame color: dark blue
    fonttitle=\bfseries\scshape,% Title font: bold small caps
    fontupper=\ttfamily,        % Set the main content font to monospace (Bera Mono)
    title={#2},                 % Set the title using the mandatory argument
    attach boxed title to top left={yshift=-0.1in, xshift=0.15in}, % Position title
    boxed title style={
        colback=yellow!75!black, % Title box background: dark blue
        colframe=yellow!75!black % Title box frame: dark blue
    },
    before upper={\vskip2mm}, % Add some space below the title box
    #1
}

\newtcolorbox{llmgreenresponse}[2][]{%
    enhanced,
    colback=green!5!white,      % Background color: very light green
    colframe=green!60!black,    % Frame color: dark green
    fonttitle=\bfseries\scshape,% Title font: bold italic
    fontupper=\ttfamily,  % Set the main content font to monospace (Bera Mono)
    title={#2},
    attach boxed title to top right={yshift=-0.1in, xshift=-0.15in}, % Position title (right)
    boxed title style={
        colback=green!60!black,
        colframe=green!60!black
    },
    before upper={\vskip2mm},
    #1
}

\newtcolorbox{llmredresponse}[2][]{%
    enhanced,
    colback=red!5!white,      % Background color: very light green
    colframe=red!60!black,    % Frame color: dark green
    fonttitle=\bfseries\scshape,% Title font: bold italic
    fontupper=\ttfamily,  % Set the main content font to monospace (Bera Mono)
    title={#2},
    attach boxed title to top right={yshift=-0.1in, xshift=-0.15in}, % Position title (right)
    boxed title style={
        colback=red!60!black,
        colframe=red!60!black
    },
    before upper={\vskip2mm},
    #1
}
\lstdefinestyle{codestyle}{
    commentstyle=\color{cardinal}\slshape,
    keywordstyle=\color{black}\bfseries,
    numberstyle=\color{codegray},
    basicstyle=\ttfamily\mdseries\scriptsize,
    emph={IC_Cache,client},
    emphstyle={\color{black}\bfseries},
    breakatwhitespace=false, 
    frame=lines,      
    rulecolor=\color{codegray},  
    breaklines=true,                 
    captionpos=b,                    
    keepspaces=true,                 
    numbers=left,                    
    numbersep=5pt,                  
    showspaces=false,                
    showstringspaces=false,
    showtabs=false,                  
    tabsize=2
}
\definecolor{cardinal}{rgb}{0.77, 0.12, 0.23}

\SetCommentSty{mycommfont}
\begin{document}
\sloppy
\title{\name: Efficient Large Language Model Serving via In-context Caching}
% \title{\fontsize{16}{16}{\textbf{\name: Efficient Large Language Model Serving via In-context Caching}}}

\author{Yifan Yu}
\authornote{Both authors contributed equally to this research.}
\authornote{Part of the work was done when working at Google.}
% \email{yifanyu4@illinois.edu}
\affiliation{%
  \institution{\textit{University of Illinois Urbana-Champaign}}
  \city{}
  \state{}
  \country{}
}
% \hspace{-1cm}
\author{Yu Gan}
\authornotemark[1]
% \email{gany@google.com}
\affiliation{%
  \institution{\textit{Google}}
  \city{}
  \state{}
  \country{}
}

\author{Nikhil Sarda}
% \email{nikhilsarda@google.com}
\affiliation{%
  \institution{\textit{Google}}
  \city{}
  \state{}
  \country{}
}

\author{Lillian Tsai}
% \email{tslilyai@google.com}
\affiliation{%
  \institution{\textit{Google}}
  \city{}
  \state{}
  \country{}
}

\author{Jiaming Shen}
% \email{jmshen@google.com}
\affiliation{%
  \institution{Google}
  \city{}
  \state{}
  \country{}
}

\author{Yanqi Zhou}
% \email{yanqiz@google.com}
\affiliation{%
  \institution{Google}
  \city{}
  \state{}
  \country{}
}

\author{Arvind Krishnamurthy}
% \email{arvindkrish@google.com}
\affiliation{%
  \institution{\textit{Google \& University of Washington}}
  \city{}
  \state{}
  \country{}
}

\author{Fan Lai}
% \email{fanlai@illinois.edu}
\authornotemark[2]
\affiliation{%
  \institution{\textit{University of Illinois Urbana-Champaign}}
  \city{}
  \state{}
  \country{}
}

\author{Henry M. Levy}
% \email{hanklevy@google.com}
\affiliation{%
  \institution{\textit{Google \& University of Washington}}
  \city{}
  \state{}
  \country{}
}

\author{David E. Culler}
% \email{dculler@google.com}
\affiliation{%
  \institution{\textit{Google}}
  \city{}
  \state{}
  \country{}
}

\renewcommand{\shortauthors}{Yu et al.}

\input{sections/abstract}

\begin{CCSXML}
<ccs2012>
   <concept>
    <concept_significance>500</concept_significance>
       </concept>
   <concept>
       <concept_id>10010147.10010257</concept_id>
       <concept_desc>Computing methodologies~Machine learning</concept_desc>
       
       <concept_id>10010520.10010521.10010537.10003100</concept_id>
       <concept_desc>Computer systems organization~Cloud computing</concept_desc>
       <concept_significance>500</concept_significance>
       </concept>
 </ccs2012>
\end{CCSXML}

\ccsdesc[500]{Computer systems organization~Cloud computing}
\ccsdesc[500]{Computing methodologies~Machine learning}


% \keywords{Do, Not, Us, This, Code, Put, the, Correct, Terms, for,
% \keywords{LLM serving, In-context caching}
\keywords{Large Language Models (LLMs), LLM serving, Cloud Computing, Semantic Caching, Request Routing, Load Balancing, Quality-Efficiency Tradeoff}

% \received{Apr 18 2025}
% \received[revised]{Aug 15 2025}
% \received[accepted]{July 11 2025}

\maketitle

\input{sections/introduction}

\input{sections/background}
\input{sections/overview}

\input{sections/design}
\input{sections/implementation}

\input{sections/evaluation}

\input{sections/related}

\input{sections/discussions}

\input{sections/conclusion}

\section*{Acknowledgments}
We sincerely thank Ana Klimovic for her valuable feedback while shepherding our paper. We also sincerely thank Martin Maas, Chandu Thekkath, Fatma Ozcan, Salem Haykal, and the anonymous reviewers for their feedback on earlier versions of this manuscript. This work was also supported in part by ACE, one of the seven centers in JUMP 2.0, a Semiconductor Research Corporation (SRC) program sponsored by DARPA. The work utilized the Delta system at the National Center for
Supercomputing Applications (NCSA) through allocation
CIS240236 from the ACCESS program.

\label{EndOfPaper}

{
\bibliographystyle{plain}
\bibliography{main}
}
\clearpage

\appendix
\input{sections/appendix}

%\end{sloppypar}
\end{document}

%% file: sections/abstract.tex
\begin{abstract}
Large language models (LLMs) have excelled in various applications, yet serving them at scale is challenging due to their substantial resource demands and high latency. Our real-world studies reveal that over \update{70\%} of user requests to LLMs have semantically similar counterparts, suggesting the potential for knowledge transfer among requests. However, naively caching and reusing past responses leads to \update{a} big quality drop.

In this paper, we introduce IC-Cache, a caching system that enables \emph{live LLM capability augmentation} to improve serving efficiency: 
by leveraging historical request-response pairs from larger models as in-context examples, IC-Cache empowers small LLMs to imitate and even exceed the compositional abilities (e.g., reasoning) of their larger counterparts, enabling selective offloading of requests to reduce cost and latency.
Achieving this live augmentation at scale introduces intricate trade-offs between response quality, latency, and system throughput. 
For a new request, IC-Cache efficiently selects similar, high-utility examples to prepend them to the new request's input.
At scale, it adaptively routes requests across LLMs of varying capabilities, accounting for response quality and serving loads.
IC-Cache employs a cost-aware cache replay mechanism that refines example quality offline to maximize \update{online} cache utility and efficiency. 
Evaluations on millions of realistic requests demonstrate that IC-Cache improves LLM serving throughput by 1.4--5.9x and reduces latency by 28--71\% without hurting response quality. 
\end{abstract}

%% file: sections/introduction.tex
\section{Introduction}
\label{sec:intro}

Large language models (LLMs) have achieved remarkable success in diverse applications like chatbots~\cite{characterai, claude}, code generation~\cite{amazon-code, copilot}, and math reasoning~\cite{team2023gemini}, already handling millions of daily user requests~\cite{dynamollm-isca24, dlora-osdi24}. 
While the trend towards ever-larger LLMs enhances service quality---DeepSeek-R1 has 671 billion parameters~\cite{deepseek-r1}---serving these large models is becoming prohibitively costly and operationally complex.

At the heart of these challenges lies a fundamental tradeoff between model quality, latency, and serving cost. The drive for high-quality outputs has led to the deployment of models with hundreds of billions of parameters~\cite{team2023gemini, llama3-report}. 
However, their substantial resource demands have led to soaring operational costs for service providers and high latency for users. Ensuring efficient serving is further complicated by the prevalence of bursty workloads, where request loads fluctuate dramatically even in minutes~\cite{helicon-google, patel2024splitwise}. As a result, service providers often have to rely on complex and/or overprovisioned infrastructure to maintain responsiveness~\cite{deepseek-r1, modserve-arxiv25}.

Recent advances in LLM serving systems have primarily focused on optimizing system throughput---such as improving parallelism~\cite{alpaserve-osdi23, megascale-nsdi24}, GPU utilization~\cite{zhong2024distserve, dlora-osdi24, agrawal2023sarathi, hygen-arxiv25}, and memory efficiency~\cite{vllm-sosp23, sglang-asplos}---and reducing serving latency (\eg, through better request scheduling~\cite{vtc-osdi24, parrot-osdi24, andes-arxiv24}). 
However, a complementary opportunity remains largely unexplored: \emph{exploiting the natural similarity among daily LLM requests}. 
Our analysis of four open-source user request datasets reveals that over \update{70\%} of requests have a semantically similar counterpart in past requests.
Unfortunately, simply caching historical requests and reusing their responses either yields low hit rates for exact matches or suffers from significant quality degradation in similarity-based matches, as any contextual mismatches can risk off-topic replies  (\S\ref{subsec:opportunity}).

In this paper, we introduce an \underline{I}n-\underline{C}ontext \underline{Cach}ing system, \name, that unlocks new sweet spots in the quality-latency-cost tradeoff for practical LLM serving as the emergence of long-context small LLMs (1B-10B)~\cite{team2024gemma,gemini_flash_8b,llama3-report}.  \update{It addresses these classic systems concerns, resource efficiency, load balancing, and response latency, through novel, end-to-end ML-system co-designs.}
Grounded by in-context learning theory~\cite{icl-theory-arxiv23, icl-theory-iclr25}, \name cautiously selects high-utility historical request-response pairs from large LLMs, and prepends them as in-context examples to the new request's input, guiding smaller LLMs to produce better responses.
This enables \emph{live capability augmentation} of LLMs, where small LLMs can imitate and sometimes even exceed the compositional abilities (\eg, reasoning) of larger models. As a result, low-latency, cost-efficient LLMs can adaptively offload requests from expensive counterparts without compromising response quality. 
Extending the caching architectures widely adopted in LLM deployments (\eg,  prefix caching~\cite{gemini-cache, deepseek-cache} and semantic caching~\cite{databricks-cache}), 
\name is lightweight, complementary to existing LLM serving systems, and can be integrated with a few lines of code changes (\S\ref{sec:overview}). 
% 
% Unlike traditional methods such as fine-tuning models or retrieving piecemeal factual information from documents~\cite{longrag-emnlp24}, \name enables customizable, on-the-fly capability augmentation (\S\ref{sec:eval:icc_vs_sft}).

\name addresses three key challenges to optimize the quality-latency-cost tradeoff. 
First, it must efficiently identify high-utility examples to improve model responses for voluminous daily requests~\cite{dynamollm-isca24, conserve-arxiv24}. 
Selecting examples based solely on relevance often leads to marginal gains, as it ignores both the model's intrinsic capabilities and the quality of examples.
\name employs a two-stage example selection mechanism that balances selection efficiency and quality. 
It first pre-selects a small subset of examples with high relevance to ensure scalability, then uses a lightweight proxy model to estimate their end-to-end utility. 
While adding more examples can further improve response quality---thus enabling \update{more} offloading from large models---it increases the input length, which raises latency and risks exceeding the context window of small models. 
\name considers the utility and coverage of examples to optimize offloading efficiency (\S\ref{sec:retrieval}).

Second, request offloading to smaller LLMs must account for response quality. 
Aggressively offloading requests can degrade response quality. 
Optimizing this efficiency-quality tradeoff requires accounting for request complexity, example utility, and current serving load. 
Further complicating this is the dynamic nature of real-world deployments, with shifts in request trends (\eg, hot topics), model performance (e.g., after upgrades), and bursty load patterns.
\name introduces a lightweight, bandit-based request router that jointly considers the request and selected examples to route requests to LLMs of varying capabilities.
The router adapts to instantaneous load conditions to strike an effective balance between efficiency and quality, and learns from recent requests to refine its policy in a resource- and data-efficient manner (\S\ref{sec:router}).

Third, managing the example cache under data and load dynamics requires optimizing example quality and efficiency at scale. The distribution of requests (\eg, new or hot topics) evolves over time, which in turn impacts the utility and access frequency of examples. 
To maximize overall cache utility, \name employs a cost-aware example replay mechanism that selectively refines example quality offline---e.g., by opportunistically replaying past examples and preserving those that yield the most effective responses---when the expected gains justify the replay cost. 
Over time, \name selectively retains \update{and} evicts examples to maintain a bounded cache size while respecting privacy (\S\ref{sec:manager}).

Our implementation of \name supports three popular LLM serving frameworks (\S\ref{sec:implementation}): HuggingFace Runtime~\cite{huggingface-api}, vLLM~\cite{vllm-sosp23}, and LangChain~\cite{langchain}. 
We evaluate \name across millions of realistic, open-source queries using both proprietary models (e.g., Gemini) and open-source families such as DeepSeek-R1, Qwen, Gemma, and Phi models (\S\ref{sec:eval}). Our evaluations show that \name can improve LLM serving throughput by 1.4--5.9x, reduce response latency by 28--71\%  without compromising response quality. 

Overall, we make the following contributions in this paper:
\begin{itemize}
    \item We introduce a novel approach to repurpose past requests for live LLM capability augmentation; 
    
    \item We design efficient mechanisms for example selection, request routing, and example management, unlocking better sweet spots in the cost-latency-accuracy tradeoff;
    
    \item We implement \name, demonstrating efficiency and quality improvements across millions of realistic open-source requests and different LLM families.
\end{itemize}

%% file: sections/background.tex
\section{Background and Motivation}
\label{sec:background}

We start by reviewing existing LLM serving deployments (\S\ref{sec:background_llm_serving}), highlighting the accuracy-cost-latency tradeoff they face based on our extensive real-world studies (\S\ref{subsec:challenges}). We then describe the opportunities that motivate our work (\S\ref{subsec:opportunity}).

\subsection{LLM Serving}
\label{sec:background_llm_serving}

Practical LLM deployments often employ a scheduler to orchestrate the execution of many user requests, optimizing per-request latency and overall system throughput~\cite{yu2022orca, agrawal2023sarathi, vtc-osdi24, zhong2024distserve}. Once scheduled, a highly optimized backend (e.g., vLLM~\cite{vllm-sosp23}) makes the best use of hardware resources to generate request responses over two sequential stages: (i) \emph{Prefilling Stage}: The LLM generates the first token---a token is a unit of text, such as a word, subword, or character, that the model processes to understand and generate language. The Time-To-First-Token (TTFT) herein is critical for real-time interactions;  
% Intermediate model representations of previously processed and generated tokens, known as the key-value (KV) cache, are stored to avoid recomputation; 
(ii) \emph{Decoding Stage}: the LLM iteratively generates each subsequent token based on both the input request and previously generated tokens, until reaching either a predefined token limit or an end-of-sequence token. Here, the focus shifts to optimizing the Time-Between-Tokens (TBT).
% to ensure smooth user interactions.

Beyond generation efficiency, ensuring high-quality output is equally important. A widely adopted evaluation strategy is the LLM-as-a-judge framework~\cite{zheng2023judging, llama3-report, team2023gemini}, where an expert LLM (e.g., GPT-4 or Gemini-1.5-Pro) compares model responses to assess quality. This yields metrics such as \update{(i) \emph{Win Rate}: the percentage of queries for which a model produces a higher-rated response than its counterparts; and (ii) \emph{Response Score}: it rates responses on a scale from significantly worse (e.g., -3) to significantly better (e.g., 3)~\cite{zheng2023judging}. Both approaches offer a scalable evaluation mechanism and have shown strong correlation with human preferences \cite{huang2024empirical}. }

\subsection{Challenges in LLM Deployment}
\label{subsec:challenges}

Serving LLMs can require hundreds of GPUs, with each request potentially taking many seconds to process~\cite{agrawal2023sarathi}, leading to orders-of-magnitude higher costs compared to traditional ML workloads such as image classification with ResNet. As request volumes continue to surge, this imposes significant costs on both LLM users and service providers.
% serving infrastructures.

\begin{figure}[t]
  \centering
    {
    \subfigure[Gemini on conversation task. \label{fig:quality_cost_gemini}]{\includegraphics[width=0.48\linewidth]{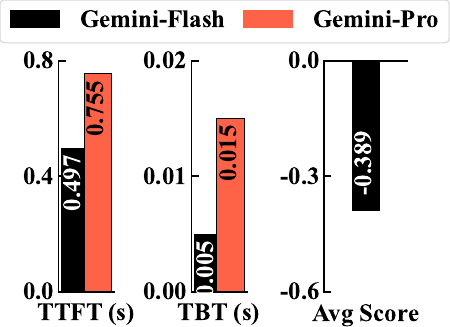}}
    \hfill
    \subfigure[
    \update{Qwen and DeepSeek.}
    % Gemma on conversation task. 
    \label{fig:direct_cache_reuse}]{\includegraphics[width=0.48\linewidth]{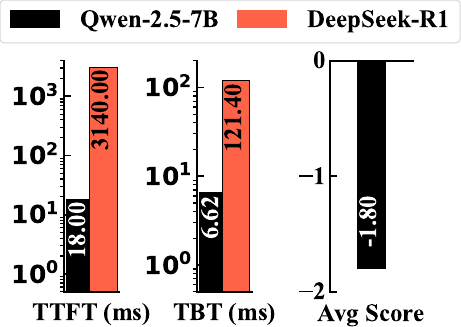} 
    }
 }
  \caption{Quality-Efficiency Trade-off of Gemini,  DeepSeek-R1, and Qwen Models. The average score represents a seven-point scale ranging from -3 (significantly worse) to 3 (significantly better), evaluating the smaller model's performance relative to the larger model. Self-comparison scores (large vs. large) are omitted as they would yield 0 by definition. 
  \vspace{-.4cm}
  }
  \label{fig:model-tradeoff}
\end{figure}

\paragraph{Accuracy-Cost-Latency Tensions in LLM Serving.}
Ever-larger LLMs continue to advance the performance frontier~\cite{arena-dashboard}, e.g., DeepSeek-R1 has 671B parameters, fueling their adoption across a wide range of applications. 
Increasingly, this cycle heightens the tension between model quality and efficiency: users expect high-quality responses and low-latency interactions, yet larger LLMs deliver high-quality results at higher expense (e.g., \update{a 10$\times$ per-token API cost difference between ChatGPT-4o and ChatGPT-3.5}).

We further delve into this cost-latency-accuracy trade-off across both proprietary and open-source LLMs.
We evaluate performance on 10K real-world user requests from the LMSys-Chat dataset~\cite{lmsys-chat}, a free-form conversational generation benchmark. 
Our experiments compare \update{Gemini-1.5-Pro and Gemini-1.5-Flash} on the proprietary side, and Qwen2.5-7B and DeepSeek-R1 on the open-source side. To ensure fair and cross-validated quality assessment, we use DeepSeek-R1 as the autorater for Gemini experiments, and Gemini-1.5-Pro as the autorater for DeepSeek-R1 experiments.
% 
%We also study the LLaMA-3B and LLaMA-70B models on a code generation task using the WMT-16 translation  benchmark~\cite{ceil-icml23} with golden labels.
% 
Figure~\ref{fig:model-tradeoff} shows that more complex and larger models produce higher-quality outputs. For example, Gemini-1.5-Pro achieves a 0.39 higher response score compared to the Gemini-Flash model, corresponding to a 65\% win rate in our win rate evaluation. However, this quality improvement comes at the cost of 3$\times$ higher TBT latency and a significantly larger resource footprint. For comparison, deploying DeepSeek-R1 requires 16 A100 GPUs, whereas Qwen-7B can run on a single GPU.

\begin{figure}[t]
  \centering
    { 
    \subfigure[Request Density over Time. \label{fig:train-ne-xs}]{\includegraphics[width=0.49\linewidth]{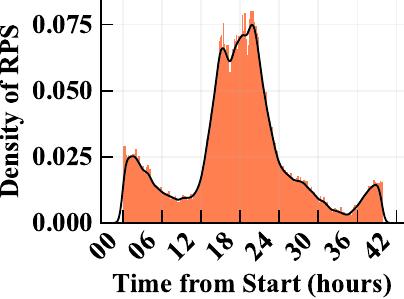}}
    \hfill
    \subfigure[Minute-level request arrivals. \label{fig:eval-ne-xs}]{\includegraphics[width=0.49\linewidth]{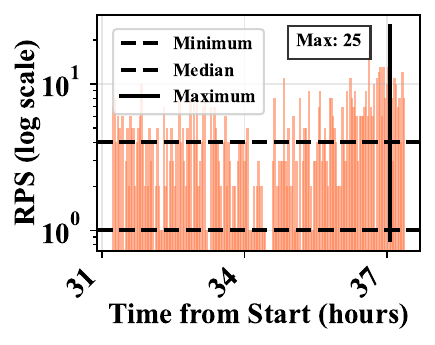}} 
  }
  \caption{Serving loads vary significantly between peak and off-peak hours (a), and even within minutes (b).
  }
  \vspace{-.3cm}
  \label{fig:load-trace}
\end{figure}

\paragraph{Substantial Scaling Demands.}
Optimizing such trade-off between accuracy, quality, and latency becomes increasingly difficult in the face of stringent service level objectives (SLOs)~\cite{adaserve-arxiv25}, growing request volumes~\cite{deepseek-r1}, and highly dynamic serving loads. 
Figure~\ref{fig:load-trace} reports our analysis of the Azure LLM serving trace~\cite{dynamollm-isca24}. 
It reveals that, beyond typical diurnal patterns, serving loads can spike dramatically in minutes, with peak loads reaching up to 25$\times$ higher than off-peak loads. 
Meeting SLOs under such transient load surges substantially increases operational complexity and often necessitates significant resource overprovisioning, particularly in private clusters.
Indeed, user requests often face severe latency fluctuations even on highly optimized proprietary LLM service platforms~\cite{helicon-google}, making SLO guarantees challenging.

\subsection{Opportunities for Repurposing Historical Requests}
\label{subsec:opportunity}

Recent advances have made strides in optimizing resource efficiency (e.g., through disaggregated resource allocation~\cite{zhong2024distserve, agrawal2023sarathi}) and user-perceived latency (e.g., via better request scheduling~\cite{vtc-osdi24, yu2022orca, hygen-arxiv25}) for individual requests and LLMs. 
Instead, we explore a complementary opportunity: leveraging historical request-response pairs as in-context examples to enable \emph{live LLM capability augmentation} at scale.
This strategy empowers smaller LLMs to produce higher-quality responses by imitating the behavior of larger models, thereby adaptively offloading traffic from them. The result is a system that significantly reduces serving costs and latency. Our approach builds on the following key observations.

\begin{figure}[t]
  \centering
    { 
    \subfigure[Query similarity of three real-world datasets. \label{fig:similarity_ms}]{\includegraphics[width=0.48\linewidth]{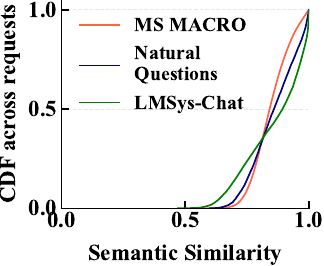}}
    \hfill
    \subfigure[Naive semantic caching based on request similarities hurts quality. \label{fig:direct_cache_hit}]{\includegraphics[width=0.48\linewidth]{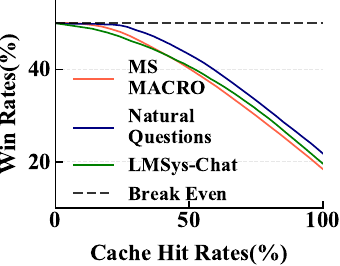}} 
    
  }
  \caption{Pervasive request similarity in the requests. We measure the top-1 request similarities of each request to others on MS MACRO~\cite{msmarco}, Natural Questions~\cite{DBLP:journals/corr/NguyenRSGTMD16}, and LMSys-Chat~\cite{lmsys-chat} datasets (Table~\ref{tab:data-stats}). Semantic caching hurts quality significantly. 
 }
  \vspace{-.3cm}
  \label{fig:query-similarity}
\end{figure}

\paragraph{Voluminous requests lead to pervasive similarity.} 
With the rapid growth in daily LLM requests, semantic similarities among them are also increasing.
We analyze the request similarity in three real-world datasets: MS MACRO (Bing Search requests)~\cite{msmarco} (1 million requests), Natural Questions (300K requests), and LMSys-Chat~\cite{lmsys-chat} (about 1 million requests). We extract the dense embedding (\ie, the model's semantic representation of text) of each request input using the popular T5-model~\cite{2020t5} and measure their cosine similarity ($\in$ [0, 1]), where 1 indicates identical requests. Figure~\ref{fig:similarity_ms} shows that more than 70\% of the requests have at least one other request with a cosine similarity greater than 0.8, a threshold commonly considered to reflect strong semantic overlap~\cite{semantic-nips18}, compared to the 0.5 similarity of random request pairs.

\paragraph{Semantic caching is widely adopted yet limits efficiency and quality.}
The abundance of similar requests in online LLM serving presents significant opportunities to reuse historical responses. Major providers---including Google Gemini, OpenAI, and DeepSeek---already support context caching mechanisms to reduce deployment costs, such as reusing KV caches for shared input prefixes~\cite{sglang-asplos}. Building on this, GPTCache~\cite{bang2023gptcache} and Databricks~\cite{databricks-cache} introduce semantic caching, which reuses responses when a new request is semantically similar to a previous one.

Despite established infrastructure support, existing semantic caching approaches remain limited in both efficiency and quality. Exact match rates are typically low due to diverse phrasings of similar queries. However, reusing semantically similar queries risks degrading output quality, as determining whether two requests are semantically equivalent is inherently subjective~\cite{semantic-nips18}. As shown in Figure~\ref{fig:direct_cache_hit}, if we select the most similar request and return its cached response, the win rate compared to generating a response using the same model drops from 50\% to 18\%.

\begin{figure}[t]
  \centering
    { 
    \subfigure[Response quality with examples. \label{fig:accuracy_qwen}]{\includegraphics[width=0.49\linewidth]{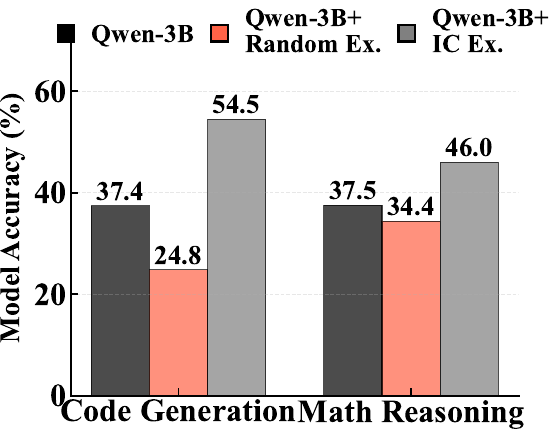}}
    \hfill
    \subfigure[TTFT performance. \label{fig:ttft_qwen}]{\includegraphics[width=0.49\linewidth]{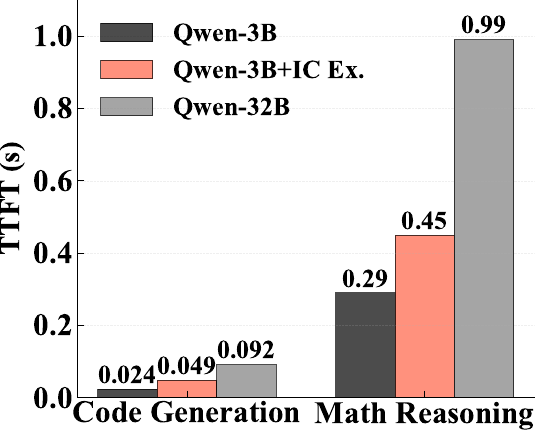}} 
    
  }
  \caption{(a) Incorporating  in-context (IC) examples improves response quality, whereas adding random examples degrades it. (b) Prepending examples introduces a slight TTFT overhead, but it remains lower than that of querying larger models.}
   \vspace{-.3cm}
  \label{fig:example-impact}
\end{figure}

\paragraph{History can enable live LLM capability augmentation.} 
Rather than risking off-topic responses, we argue that historical request-response pairs can be repurposed as in-context examples, prepended to new requests to enable live capability augmentation for smaller models.
This insight is grounded in in-context learning theory~\cite{icl-survey}, which demonstrates that LLMs can learn from high-quality examples, enabling on-the-fly knowledge transfer and skill imitation. 
This augmentation allows smaller models to mirror not only surface-level answer patterns (e.g., structure or detail) but also the reasoning trajectories of larger models. In contrast, Retrieval-Augmented Generation (RAG), a different approach for adding additional information to the context, performs piecemeal factual knowledge lookups over static documents, lacking the compositional reasoning captured in LLM responses. 

We validate this opportunity for augmentation by comparing Qwen2.5-3B augmented with five examples from Qwen2.5-32B on the NL2Bash code generation~\cite{ye2023generating} and Math-500-Hard reasoning benchmarks~\cite{dpsynth}. Our extensive evaluations on various free-form generation tasks report consistent \update{results} (\S\ref{eval:e2e}). 
Figure~\ref{fig:accuracy_qwen} shows that small LLMs with well-selected in-context examples can significantly improve response quality, while random examples degrade performance due to distractions. 
Figure~\ref{fig:ttft_qwen} shows that prepending examples slightly increases prefilling time due to longer input---the decoding time remains largely unaffected---but still far lower than that of large models.

\paragraph{Design Challenges.}
As online LLM serving \update{deployments} routinely process large volumes of requests, they provide a continuous stream of fresh, in-distribution examples. This enables customized, live augmentation of model capabilities without retraining. However, doing so introduces unique challenges:
\begin{itemize}
\item \emph{Accuracy}:
Examples are collected online, introducing variability in quality and diversity. Selecting high-utility examples requires going beyond relevance; it must account for example quality, diversity, and the capabilities of the target model to maximize final response quality. Further, how to ensure response quality when offloading requests between LLMs of varying capabilities?

\item \emph{Efficiency}:
How can we efficiently select examples and manage their cache across voluminous daily requests? 
How to react to fluctuating serving loads and evolving data distributions, optimizing for the optimal accuracy-latency-cost trade-off in flight?
\end{itemize}

%% file: sections/overview.tex
\section{\name Overview}
\label{sec:overview}

Extending the prevalent semantic caching architecture in today's LLM deployments, we introduce \name, an in-context caching LLM serving system.
It exploits historical requests for live LLM capability augmentation, adaptively offloading requests to reduce serving latency and cost while maintaining response quality.

\paragraph{Design Space and Architectures.} 
\name benefits today's LLM serving paradigms across three popular settings: 
(i) \emph{Cloud Deployment}:
\name enables the offloading of requests from larger LLMs to smaller, more efficient models, providing users with responsive, high-quality outputs while reducing service providers' deployment costs;
(ii) \emph{Edge Deployment}:
By personalizing the selection of historical examples, \name empowers small, on-device LLMs (e.g., Apple Intelligence~\cite{apple-ai} and Copilot + PC~\cite{copilot}) to generate higher-quality responses, alleviating the costs  associated with cloud-based solutions;
(iii) \emph{Edge-Cloud Deployment}:
In collaborative edge-cloud scenarios, \name enables smaller models to generate better responses locally while selectively routing requests to the cloud. 
In this paper, we focus on online LLM serving in the cloud and demonstrate IC-Cache's effectiveness across various deployment scenarios (\S\ref{sec:eval}). 

% \paragraph{System Architecture.}
\begin{figure}[t]
  \centering
  \includegraphics[width=0.99\linewidth]{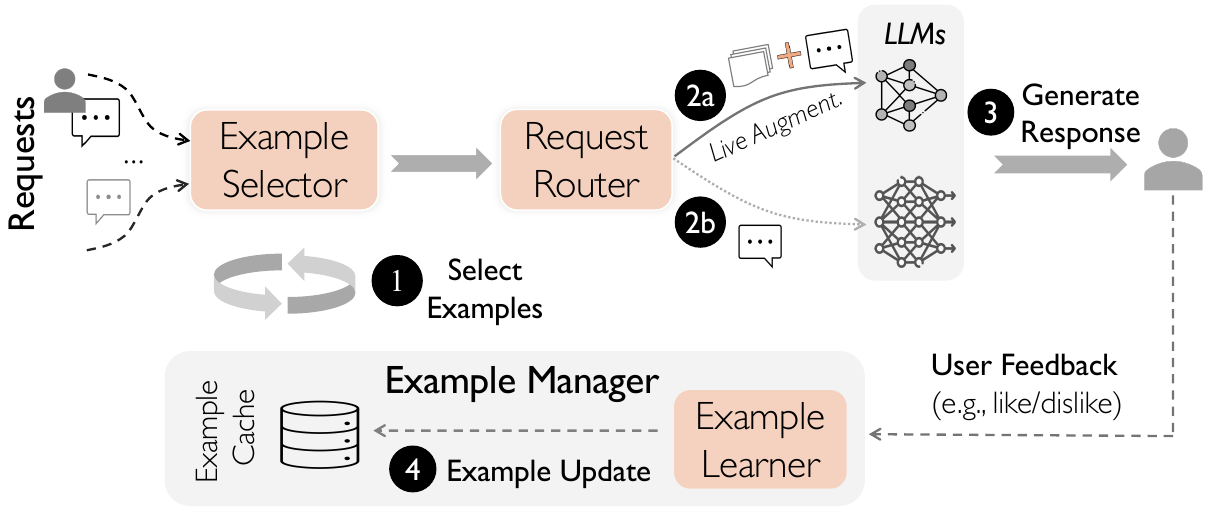}
  \caption{\name overview and request execution flow. }
  \vspace{-.3cm}
  \label{fig:sys-arc}
\end{figure}

As shown in Figure~\ref{fig:sys-arc}, \name serves as a complementary layer between existing serving systems (\eg, vLLM~\cite{vllm-sosp23}) and LLM applications. It comprises three key system components:
\begin{itemize}
\item \emph{Example Selector}: 
It selects high-utility request-response pairs from history as examples to augment LLMs. 

\item \emph{Request Router}: 
It routes new requests to the most suitable LLM, such as small models augmented with examples, or larger models, based on request complexity and the current serving load. 

\item \emph{Example Manager}:
It manages the caching of requests over time (\eg, eviction) and opportunistically improves example quality (\eg, by asynchronously replaying requests and storing the best response).
\end{itemize}

\begin{figure}[t]
\begin{lstlisting}[xleftmargin=3.5ex,language=Python,label={lst:keeper_client},escapechar=|]
from IC_cache import IC_cacheClient

def generate_response(requests, generation_config):
    # Create client session to IC-Cache service
    client = IC_cacheClient(config=generation_config)

    # Get inference results with generation config 
    response = client.generate(requests)

    # Register requests to the cache  
    client.update_cache(requests, response)
    client.stop()
\end{lstlisting}
  \caption{\name benefits LLM serving with a few lines of code.}
  \vspace{-.3cm}
  \label{code:sys-agent}
\end{figure}

\begin{figure*}[t]
  \centering
    \begin{minipage}{0.25\linewidth}
    \centering
    \includegraphics[width=1.0\linewidth]{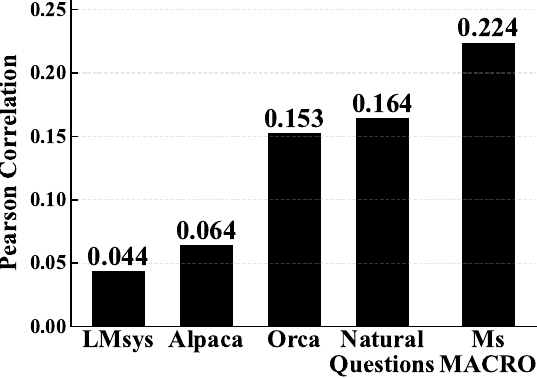}
    \caption{Pearson correlation ($\in$ [-1, 1]) between example similarity and its helpfulness is weak. }
    \label{fig:sim-utility}
  \end{minipage}
  \hfill
  \begin{minipage}{0.49\linewidth}
  \centering
 \includegraphics[width = 0.9\linewidth]{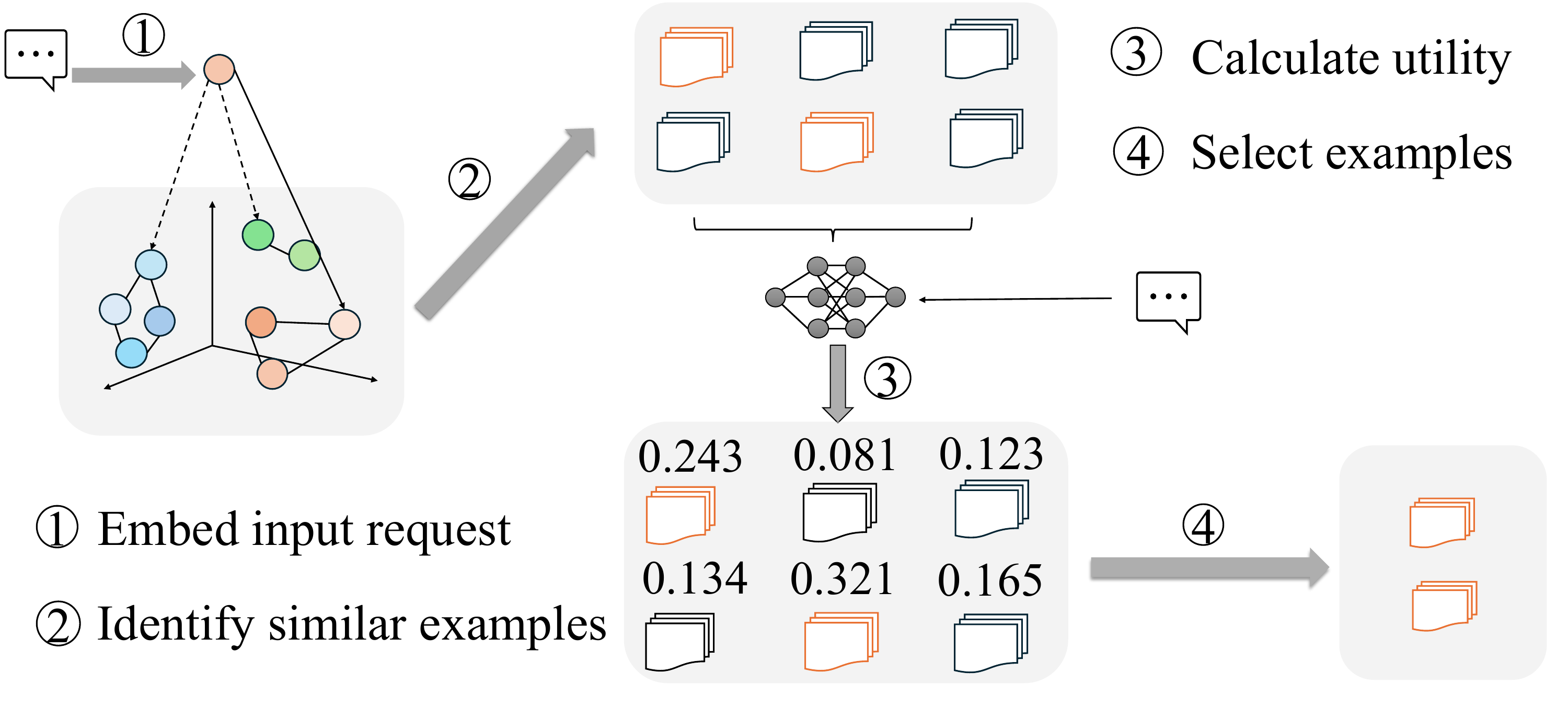}
  \caption{Overview of two-stage example selection. }
  \label{fig:example_selection}
\end{minipage}
 \hfill
  \begin{minipage}{0.25\linewidth}
    \centering
    \includegraphics[width=1.0\linewidth]{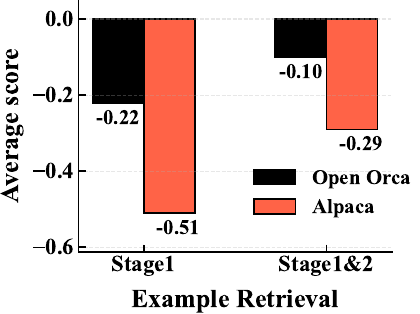}
    \caption{Two-stage example selection improves response quality.}
    \label{fig:example_select_quality}
  \end{minipage}
%   \hfill

\end{figure*}

\paragraph{Serving Workflow}
Figure~\ref{code:sys-agent} illustrates the \name user interface, complementing existing serving paradigms with minimal code modifications. 
Upon receiving new requests from users  (Figure~\ref{fig:sys-arc}):
\blackcircled{1}
The Example Retriever retrieves the most helpful request-response pairs (e.g., based on relevance and quality) from the cache to serve as in-context examples.
\blackcircled{2}
The new request, now augmented with these examples, is passed to the Request Router, which determines the appropriate LLM to handle the request.
\blackcircled{3}
The selected model processes the request and generates a response, following the usual LLM generation process. The response is then delivered to the user.
\blackcircled{4}
Finally, the Example Manager may add the request-response pair to the cache, depending on application-specific requirements (e.g., removing sensitive information), improve example quality via cost-aware replay, and evict stale or low-quality entries over time.

%% file: sections/design.tex
\section{\name Design}
\label{sec:design}

In this section, we describe how \name unlocks new Pareto frontiers for accuracy-efficiency trade-offs in existing LLM serving, by selecting helpful examples at scale (\S\ref{sec:retrieval}), adaptively offloading requests among LLMs (\S\ref{sec:router}), and managing examples to improve efficiency and utility (\S\ref{sec:manager}).

\subsection{Example Selector: Select High-Utility Examples}
\label{sec:retrieval}

The utility of an example is defined by its effectiveness in improving response quality. However, directly measuring this utility post hoc is impractical, as it requires generating model responses conditioned on each example, a costly and time-consuming process. A common alternative is to approximate utility using semantic relevance, such as the cosine similarity between text embeddings, following practices from RAG systems~\cite{databricks-cache}. 
However, as shown in Figure~\ref{fig:sim-utility}, semantic relevance exhibits only a weak correlation with true example helpfulness, limiting its reliability as a utility proxy.

This limitation arises because relevance-based selection fails to account for model-specific capabilities and example quality, resulting in biased utility estimation. For instance, examples containing low-quality responses or covering skills the smaller model already handles well contribute little to quality improvement, and may even degrade generation quality while incurring additional overhead (Figure~\ref{fig:example-impact}). 
Furthermore, while relevance can enrich response details, overall response quality depends on a broader set of dimensions, such as accuracy, depth, and creativity~\cite{zheng2023judging}, which extend far beyond relevance.

Next, we present a two-stage hierarchical example selection mechanism to select high-utility examples at scale, and extend it to select the combination of examples.

\paragraph{Two-Stage Scalable Example Selection.}
Our key insight is that, while semantic relevance alone has a weak correlation with actual helpfulness, it remains a useful lightweight filter (Figure~\ref{fig:sim-utility}).
We exploit this by first applying relevance-based selection to narrow down the candidate pool. 
Since directly estimating the helpfulness of each example through rule-based heuristics is impractical, we introduce a second-stage proxy model to estimate the fine-grained helpfulness of examples on a per-request basis. This two-stage selection design enables request- and model-aware example retrieval while avoiding the prohibitive cost of scanning the entire example pool.

As illustrated in Figure~\ref{fig:example_selection}, when a new request arrives, we compute its semantic similarity to cached examples using dense text embeddings and prioritize examples with high similarity scores.
To ensure scalability, we can cluster cached examples offline into $K$ groups using K-Means and process incoming requests online. 
However, the choice of $K$ introduces a trade-off: using too many clusters increases the cost of identifying the nearest centroid (i.e., $K$ comparisons), while too few clusters increases the cost of searching within clusters (i.e., $N/K$ comparisons per cluster on average). To minimize total matching cost per request, we balance these components by solving: $K = \arg\min_K \left(K + \frac{N}{K}\right)$, yielding $K = \sqrt{N}$.

From the preselected candidate pool, we apply a lightweight proxy model to estimate each example's helpfulness. The model takes as input the current request and a candidate request-response pair, estimating the example's end-to-end helpfulness to \update{improve} the final response.
Our design builds on existing infrastructure in practical LLM serving platforms, where user feedback, such as thumbs-up/down ratings and preference comparisons (e.g., "Which response do you prefer?" in ChatGPT, Gemini, and DeepSeek), is often collected. 
Moreover, serving systems often sample and evaluate response quality using reward models or human feedback to monitor serving performance over time~\cite{ouyang2022training,dai2023safe}. 
Given the scale of modern LLM deployments, even a small sampling rate (e.g., 1\%) across millions of daily requests~\cite{deepseek-r1} already yields tens of thousands of feedback, sufficient to continuously update the proxy model offline. 
We note that the proxy model is lightweight (\eg, TinyBert has <0.2\% model size of Qwen2.5-7B),  updated asynchronously.

Figure~\ref{fig:example_select_quality} shows \update{that} the two-stage mechanism greatly improves response quality with little (<1\%) overhead  (\S\ref{eval:breakdown}).

\paragraph{Selecting Example Combinations.}
While our two-stage selection mechanism identifies high-utility examples, including too many yields diminishing quality improvements, implying marginal efficiency gains from offloading. Moreover, longer input sequences increase inference costs after offloading (Figure~\ref{fig:example-impact}), reducing the net benefit of using examples. Thus, the number of examples should be adapted per query.
% to balance quality and efficiency.

Our key insight is that LLM serving is a long-running process, which enables \name to explore and adapt to the impact of varying example counts continuously over time.
Building on the two-stage selection process, the Example Selector uses a dynamic utility threshold to filter out low-impact examples, excluding those whose estimated helpfulness falls below the current threshold.
During online deployment, \name periodically samples a subset of requests and evaluates the average efficiency gains achieved under different utility thresholds (e.g., utility > 0.5). It then selects the threshold that maximizes overall performance and applies it globally. 
Note that our Request Router will ensure the response quality of sampled requests is preserved (\S\ref{sec:router}), making this adaptive process robust. 
This ensures that the number of selected examples is both query- and example-dependent, continuously optimized to maximize end-to-end efficiency.

\subsection{Request Router: Trade off Efficiency and Accuracy}
\label{sec:router}

\begin{algorithm}[t]
\caption{Pseudo-code of \name runtime}\label{alg:example_selection}
\label{code:pseudo_code_example}

\SetKwFunction{FSelectExamples}{\textrm{RetrieveExamples}}
\SetKwFunction{FRequestServing}{\textrm{ServeRequests}}
\SetKwFunction{FManageExamples}{\textrm{ManageExamples}}

\SetKwProg{Fn}{Function}{:}{}
\Fn{\FRequestServing{$request$, $sys\_load$}}{
    \tcc{Example Retriever: select an example combination in terms of helpfulness to improve response quality.}
    $examples \gets$ RetrieveExamples(request)
    
    \BlankLine
    \tcc{Request Router: select a model based on system load and expected response quality. Prepend examples to the request if offloading occurs.}
    $model \gets$ RouteRequest($request$, $examples$, $models$, $sys\_load$) \label{alg:router-start}
    
    $response \gets$ GenerateResponse($model$, $request$, $examples$[Optional]) \label{alg:router-end}
    
    \BlankLine
    \tcc{Example Manager: selectively cache requests in example pool and optimize example quality.}
    ManageExamples($request$, $response$)
    
    \BlankLine
    \Return $response$
}

\vspace{2mm}
\Fn{\FSelectExamples{$request$}}{
    \tcc{Stage 1: Lightweight relevance selection} \label{alg:retrieve-start}
    $request\_emb \gets$ ExtractEmbedding($request$)
    
    $relevant\_examples \gets$ GetSimilarExamples($request\_emb$, $examples$)
    % $candidates \gets$ GetExamplesFromCluster($closest\_cluster$)\;
    
    \tcc{Stage 2: Helpfulness prediction}
    \For{$ex$ in $relevant\_examples$}{
        $helpfulness \gets$ PredHelpfulness($request$, $ex$)
    }
    
    \BlankLine
    \tcc{Optimize example combination by accounting for example diversity and ordering}
    $selected\_examples \gets$  RetrieveComb($relevant\_examples$, $helpfulness$)

    \BlankLine
    \Return $selected\_examples$ \label{alg:retrieve-end}
}

\end{algorithm}
% \vspace{-.3cm}

With carefully selected examples, small LLMs gain live capability augmentation and can generate high-quality responses, enabling them to handle requests that would otherwise require larger, more expensive models. However, aggressively offloading to small LLMs can hurt response quality as the augmentation can be limited; Being overly conservative limits efficiency gains. 
While one could frame request routing as a classification problem (e.g., using a BERT-based model to predict the optimal model per request~\cite{routellm}), this approach introduces significant systems challenges, since an effective request router must: 
(i) respond to dynamic serving loads and changing quality-throughput tradeoffs; 
(ii) adapt efficiently to evolving data and model characteristics, including shifts in request distributions, example utility, model weights, and even available model sets; 
and (iii) remain execution- and data-efficient, avoiding the prohibitive cost of generating and labeling responses across all model choices.
These requirements render heavyweight classifier-based routers impractical.

To address these practical challenges, 
we propose modeling the routing decision as a contextual multi-armed bandit (MAB) problem, a lightweight and data-efficient approach often used in online recommendation systems. 
Here, the context (input) includes the request's question and its selected examples, while each ``arm'' corresponds to a candidate model (e.g., a small LLM with examples or a large LLM without). 
MAB explores and exploits by pulling different arms across trials (e.g., user requests), aiming to maximize cumulative rewards such as maximizing response quality.
Its lightweight nature and minimal needs for online feedback make it especially well-suited for real-time serving~\cite{mab-www10}.

Algorithm \ref{code:pseudo_code_example} illustrates how \name efficiently finds the sweet spot between response quality and system efficiency under dynamics. 
After the Example Selector identifies high-utility examples (Lines \ref{alg:retrieve-start}-\ref{alg:retrieve-end}), \name invokes our MAB-based Request Router to select the most suitable model for that request. The router continuously updates its offloading policy with user feedback, factoring in both perceived response quality and current system load, allowing it to adapt its routing decisions on the fly (Lines \ref{alg:router-start}-\ref{alg:router-end}).

\paragraph{Load- and Quality-aware Request Offloading.}
To handle load fluctuations,
the Request Router incorporates a load-aware biasing strategy. 
Specifically, it tracks the Exponential Moving Average (EMA) of \update{the} system serving load over time. 
When the EMA remains below the desired operational threshold (e.g., the service capacity of large models), the router prioritizes response quality.
In this regime, many requests may still be offloaded to small models, \name enables small LLMs to match or even outperform larger models on a substantial fraction of requests (\S\ref{eval:e2e}), much like how primary school teachers may better engage young learners than university professors after the right pedagogical augmentation.

In contrast, when the EMA exceeds the operational threshold, the router triggers a feedback controller to compute a corrective bias. 
This bias is calculated using the hyperbolic tangent (tanh) function~\cite{tanh-nips23}, applied to the positive load deviation (i.e., current load $-$ threshold). The resulting bias adjusts the bandit's output logits, reducing the selection scores of high-cost models and favoring more efficient, lower-cost alternatives to relieve system pressure. 
%% Fan: ptal at https://www.mathworks.com/help/slcontrol/ug/design-sliding-mode-control-reaching-law.html for the underlying control theory logic for this approach.
% 
This design offers several advantages: the tanh function provides a smooth, saturating response, enhancing stability by preventing unbounded bias values, and the bias is only active during actual overload conditions. Crucially, this lightweight control mechanism adjusts routing preferences without modifying or retraining the underlying Request Router, effectively decoupling overload management from the core routing logic. Importantly, the persistent magnitude of this applied bias can be used as a signal for infrastructure auto-scaling.

\paragraph{Efficient Router Adaptation under Dynamics.} 
In practical deployments, the Request Router must adapt to evolving data distributions and shifts in model behavior (e.g., due to model upgrades).
While retraining the router is computationally inexpensive, thanks to our contextual bandit model's compact size (0.5 million parameters), the primary challenge lies in acquiring sufficient feedback for adaptation, which may be delayed, sparse, or costly, especially when relying on explicit user signals like thumbs-up/down ratings.
To address this, we design a cost-efficient feedback collection mechanism that enables effective adaptation with minimal data overhead.
The key insight is to solicit feedback selectively, focusing only on requests where the router exhibits high uncertainty in its decision. 
We quantify uncertainty using the model's output confidence scores, which reflect how decisively the router ranks candidate LLMs.
Only requests with near-uniform confidence distributions (\eg, with standard deviation below 0.1) are tagged for feedback solicitation.
For these uncertain cases, we always include the top-ranked LLM, and probabilistically sample a second choice based on its relative confidence, encouraging adequate exploration.
Following current interface features (e.g., ``Which response do you prefer?'' in ChatGPT, Gemini, and DeepSeek), \name collects user preference feedback to refine its routing model.
We formally analyze the sample complexity of this selective feedback strategy and demonstrate its cost-effectiveness in Appendix~\ref{app:sample-complexity}.

Our evaluations show that our bandit-based router scales effectively, outperforming state-of-the-art alternatives and adapting to dynamics (\S\ref{eval:breakdown}).

\subsection{Example Manager: Effective Caching in the Wild}
\label{sec:manager}

Maximizing end-to-end efficiency hinges on (i) improving the utility of individual examples to enable more effective per-request offloading, and subsequently (ii) maintaining a high-utility example pool for offloading various requests under capacity and privacy constraints. 
Yet, real-world LLM serving systems process millions of requests daily~\cite{dynamollm-isca24, deepseek-r1}, with dynamic changes in data distribution and model behavior (e.g., trending topics), making example management a non-trivial challenge. \name includes an Example Manager that curates and optimizes examples over time. 

\paragraph{Cost-aware Example Replay.}
Recent LLM advances~\cite{scaling-law-inf, best-of-n} reveal large variance in response quality due to the stochastic nature of generation (e.g., token sampling). This variance can be harnessed through example replay, where the same request is queried multiple times, and the best response is selected for reuse. 
By refining example responses in this way, we can boost their downstream utility for offloading. 
This example replay can be performed offline (e.g., during off-peak hours) to avoid runtime latency. Understandably, the relative cost can be negligible, considering that a refined example reused hundreds of times incurs only around 1\% amortized overhead. 
But serving at scale, it is still important to curb the resource overhead from generating new responses. 

\begin{figure}[t]
  \centering
  \begin{minipage}[t]{0.48\linewidth}
    \centering
    \includegraphics[width=\linewidth]{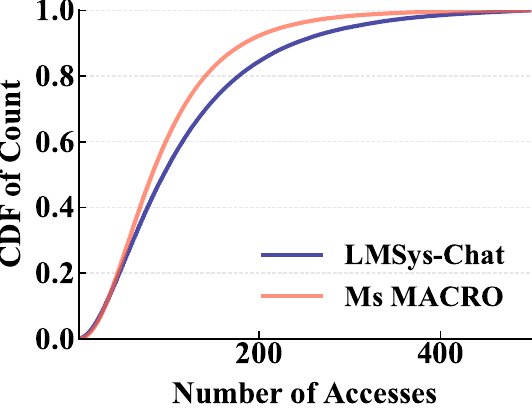}
    \caption{Example access exhibits long-tail distribution.}
    \label{fig:example_access_feq}
  \end{minipage}
  \hfill
  \begin{minipage}[t]{0.5\linewidth}
    \centering
    \includegraphics[width=\linewidth]{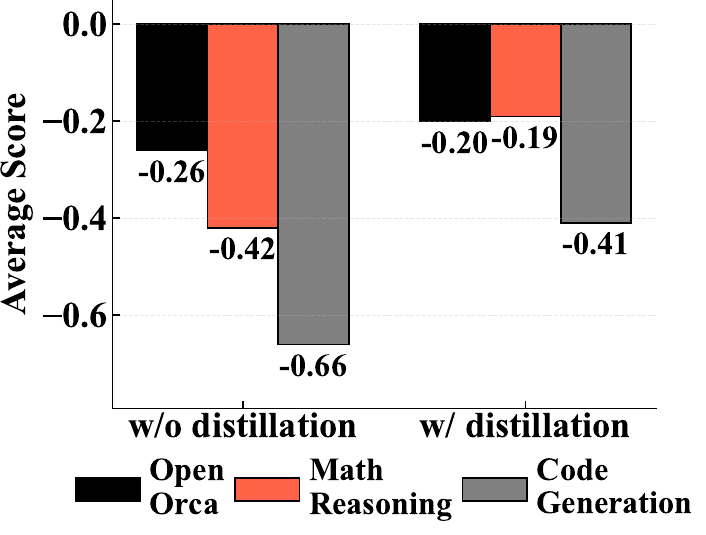}
    \caption{Example distillation improves final response quality.}
    \label{fig:example_distillation}
  \end{minipage}
\end{figure}

In practice, both the frequency and effectiveness of example reuse exhibit large heterogeneity, often following a long-tail distribution (Figure~\ref{fig:example_access_feq}). 
Hence, the Example Manager introduces a cost-aware example replay mechanism that selectively replays examples based on their potential efficiency gains. 
Intuitively, when repurposing an example, the smaller the model to which its augmented request can be routed and the higher the response quality it achieves, the smaller the gain we can expect from further refining the example's response. 
So we can define the potential gain of improving an example $e$ as: $G(e) = (1 - normalized\_response\_quality) \times normalized\_model\_cost$. \footnote{This multiplicative formulation captures the potential quality improvement per unit cost, aligning with our objective of maximizing cost-efficiency.}
Here, $normalized\_response\_quality$ reflects user feedback (e.g., thumbs up/down) on the response quality of the example-augmented request, while $normalized\_model\_cost$ denotes the relative cost (\eg, API pricing) of the model used to serve the request. 
Since both metrics are readily available in production deployments, computing $G(e)$ is lightweight and practical. Each time $e$ is repurposed, its $G(e)$ accumulates, and \name maintains an exponentially moving average of $G(e)$ to account for evolving usage patterns.
As a result, examples that are frequently repurposed, or with them, the new requests still require larger models, or yield lower-quality responses, are prioritized for replay.

% \begin{figure}[t]
%   \centering
%     { 
%     \subfigure[Example Expansion \label{fig:example_expansion}]{\includegraphics[width=0.5\linewidth]{Figures/design/expansion.pdf}}
%     \hfill
%     \subfigure[Example Distillation \label{fig:example_distillation}]{\includegraphics[width=0.44\linewidth]{Figures/design/example_distillation_score.pdf}} 
%   }
%   \caption{Example distillation and expansion improves example quality and coverage, thereby optimizing response quality and system efficiency by enabling more effective request offloading. \todo{(b) add more tasks. replace (a) with e2e efficiency breakdown?} \fan{DS. Qwen. (b) translate into speedup.}}
%   \label{fig:expansion-distillation}
% \end{figure}

With the above formulation of potential replay gain, the Example Manager selects examples to replay whose potential resource savings outweigh the cost of replay itself. Specifically, it ranks examples by their potential gain, $G(e)$, and stops replaying examples whose potential gain falls below a cut-off. This cut-off is determined online, by the point at which replaying a higher-ranked example no longer leads to additional offloading, meaning its resource savings are lower than the one-time replay (generation) cost. 

Figure~\ref{fig:example_distillation} demonstrates that our example replay design achieves a better response quality for new requests on the Gemini-Flash model compared to the Gemini-Pro, suggesting efficiency gains during online serving.

% \paragraph{Capacity-aware Example Eviction}

% Handling a large number of daily requests requires carefully balancing memory efficiency with online computation efficiency when deciding how to cache examples. Specifically, there are two caching options:
% (i) \emph{Plaintext Request-Response Pair}: 
% This memory-efficient approach stores examples in their plaintext format. However, when prepending the example to an input during serving, it requires recomputing the KV cache from scratch, resulting in higher computation overhead during runtime; and 
% (ii) \emph{KV Cache (Model Representation)}: 
% Recent advances~\cite{promptcache-mlsys24} demonstrate that caching the precomputed KV cache of stationary inputs and concatenating them during serving can significantly reduce the prefilling overhead, with  techniques like CacheBlend~\cite{yao2024cacheblend} safeguarding quality by allowing lightweight, selective recomputation on a small subset of tokens. 
% As shown in Figure~\ref{fig:ttft_kv_caching}, caching the KV cache substantially reduces \name's time to first token (TTFT) latency. However, this method incurs a large memory footprint as KV cache is thousands of times larger than plaintext. 

\paragraph{Capping Example Cache Size over Time.} 
The Example Manager follows the deployment of popular semantic caching designs~\cite{longrag-emnlp24, databricks-cache} to minimize both overhead and engineering complexity: caching historical requests in plaintext. Plaintext caching offers low memory consumption---storing one million LMSys-Chat examples requires only about 1GB of memory (about 1\% of an A100 GPU)---and facilitates broader reuse across different models.
In fact, due to the high similarity among requests and their long-tailed access distribution, our evaluation on millions of real-world queries shows that caching just tens of thousands of examples already yields diminishing returns (\S\ref{eval:ablation}). Hence, \name introduces small memory footprints. 

% The Example Manager follows the design of existing semantic caching systems to minimize additional overhead and engineering efforts~\cite{longrag-emnlp24, databricks-cache}, which caches historical requests in plaintext for small memory footprint---caching the plaintext of one million LMSys-Chat examples takes approximately one gigabytes memory (~1\% memory size of an A100 GPU)---and better reuse across models. 

% cachingthe plaintext of one million LMSys-Chat examples takes ap-proximately one gigabytes
% Caching the KV cache of all examples would require an excessive amount of memory---caching the plaintext of one million LMSys-Chat examples takes approximately one gigabytes but their KV cache demands over 100 terabytes (Figure~\ref{fig:kvsize}). 

% This discrepancy motivates a hybrid approach in \name: it caches the plaintext of all examples while storing the KV cache only for a subset of  examples. 
% This design leverages the heterogeneous nature of example repurposing frequency (Figure \ref{fig:example_access_feq}) and the heterogeneous KV cache size due to their varing text length (Figure~\ref{fig:cache_distribution}).

For deployments with strict memory budgets, the Example Manager employs an online cache management policy that evicts low-utility examples. 
The decision process mirrors a classic knapsack problem: each example is treated as an item with a weight (its cache size, such as plaintext length) and a value (the achievable efficiency gain). The objective is to maximize the total value, \ie, the cumulative efficiency gains from caching the selected examples. The solution yields a binary caching decision for each example: whether to retain it in the cache or evict it. 
The efficiency gain of an example is measured by the number of successful offloadings it enables. To adapt to changing request patterns over time, we maintain a moving average of this gain, applying a decay factor of 0.9 every hour to emphasize recent usage while gradually discounting stale patterns.

This one-dimensional knapsack problem can be solved efficiently (\S\ref{sec:implementation}). Our solver runs periodically in the background or whenever the memory limit is approached, ensuring that cache optimization does not interfere with online serving. Evaluations on millions of real-world requests show that our strategy consistently improves performance, even under tight memory constraints (\S\ref{eval:ablation}).

\paragraph{How Does \name Respect Privacy?} 
\name follows a well-established practice of using information in cached queries to serve future requests (e.g., context or semantic caching in production systems like Databricks~\cite{databricks-cache}, Gemini~\cite{gemini-cache}, DeepSeek~\cite{deepseek-cache}, and open frameworks like GPTCache~\cite{bang2023gptcache}). \name builds on this foundation, advancing it toward live LLM capability augmentation and offloading.
We also observe that many LLM use cases where \name can provide large benefits involve only non-private data. For instance, public information retrieval and general knowledge Q\&A---such as those in Google's AI Overview~\cite{google-ai-overview} and ChatGPT Deep Search~\cite{chatgpt-dr}---make up a substantial LLM traffic.

To further ensure that the cache respects user privacy, \name controls sensitive query admission into the cache by (i) providing simple APIs for access control (Figure~\ref{code:sys-agent}), choosing whether to cache and allowing cached data sharing only within designated user domains (\eg, within the organization); and (ii) locally sanitizing inputs by removing sensitive data on the client endpoints before adding examples to the cache
(e.g., \name removes personally identifiable information using the widely adopted tool spaCy~\cite{spacy-github}).
%

% Furthermore, if developers know the sensitivity of certain examples, \name provides 
% simple APIs (Figure~\ref{code:sys-agent}) that allow developers to choose whether to cache or skip saving specific requests. Our evaluations (\S\ref{sec:eval}) demonstrate that \name improves both efficiency and quality without relying on sensitive information.
% \todo{Lily: How is the above? I tried to make more clear what is optional/developer-dependent and what is integrated.}
%user-friendly 
%developers to specify caching preferences (\eg, whether to register and store the requests for improving future responses). \todo{Lily: I don't quite understand this first point... is the point that a user can request that their queries are never stored?}\fan{Yes.}
%
% \todo{Lily's Question: the test querysets also don't have sensitive information though, right? So this could be a biased result compared to test evaluations where queries can be sensitive. Nikhil's response: Conversation and Q&A datasets are very similar to practical datasets from real users for useful applications like NotebookLM.}
% 

%
If developers require very strict privacy guarantees, \name offers replacing the historical example cache with a differentially private (DP) synthetic example cache~\cite{dpsynth}. DP synthesis ensures that an adversary with access to the synthetic examples cannot infer (with high probability) the presence or value of any specific example in the original dataset. 
Our empirical results demonstrate that \name, even when using DP-synthesized examples, continues to deliver substantial improvements over the baseline (\S\ref{eval:ablation}).

%% file: sections/implementation.tex
\section{Implementation}
\label{sec:implementation}

We implemented a prototype of \name to enable efficient LLM serving across GPUs. Our implementation leverages components already prevalent in modern LLM deployments, such as semantic caching systems~\cite{sglang-asplos, databricks-cache}, introducing minimal complexity. The \name prototype consists of approximately \update{3,000} lines of code \update{to support} widely used platforms, including vLLM~\cite{vllm-sosp23}, HuggingFace Runtime~\cite{huggingface-api}, and LangChain~\cite{langchain}, with a few lines of integration code (\S\ref{sec:overview}).

% We implemented a prototype of \name to enable efficient LLM serving across GPUs. Our system is designed to be complementary to existing LLM serving frameworks, supporting the widely used platforms such as vLLM~\cite{vllm-sosp23}, HuggingFace Runtime~\cite{huggingface-api}, and LangChain~\cite{langchain} in a few lines of code (\S\ref{sec:overview}).

\paragraph{\name Backend}
\name's backend supports distributed deployments across machines. The Example Retriever, Request Router, and Example Manager run in separate processes that can scale horizontally and communicate via gRPC.
The Example Retriever utilizes GPU-accelerated FAISS~\cite{faiss-gpu} for high-throughput similarity search.
The Request Router was implemented in Jax~\cite{jax2018github}. It includes the contextual bandit algorithm along with a prioritized replay buffer.
To improve robustness against outliers, the Example Manager follows existing advances~\cite{scaling-law-inf} to filter out examples in replay that have undergone more than five replay iterations.
To ensure high availability and load balancing, we maintain multiple replicas of \name components for load balancing. 
% The client agent interacts with the \name backend via TCP.
% connections. 

\paragraph{Fault Tolerance} 
\name \update{maintains} system state and metadata in a distributed manner across replicas, with each component periodically checkpointing its state. If a failed request to the Example Retriever or Request Router is detected, the system automatically bypasses these components and routes the request directly to the inference backend to maintain service continuity. Each component runs a lightweight daemon process that monitors service health and initiates automatic recovery procedures upon detecting failures.

%% file: sections/evaluation.tex
\section{Evaluation}
\label{sec:eval}

We evaluate \name on millions of publicly available, real-world user requests, using both open-source and proprietary models. The key results are summarized as follows:

\begin{itemize}
\item \name reduces serving latency by 28--71\% and improves throughput by 1.4--5.9$\times$ without comprising response quality (\S\ref{eval:e2e}).
\item \name can potentially offload all requests to smaller models, achieving sweet spots of efficiency-quality tradeoffs under bursty workloads (\S\ref{eval:e2e}-\S\ref{eval:breakdown}); 
\item \name  improves performance over a wide range of settings and outperforms its design counterparts (\S\ref{eval:ablation});
\end{itemize}

\subsection{Methodology}
\label{sec:exp-setup}

\paragraph{Experimental setup.}
We evaluate \name using proprietary Gemini-1.5-Pro and Gemini-1.5-Flash models via Google Cloud Vertex AI APIs, and open-source models, including DeepSeek-R1, Qwen2.5, Gemma-2, and Phi-3 model families. 
% Model specifications are summarized in Table~\ref{tab:model-sizes}. 
We summarize the datasets used in our evaluations in Table~\ref{tab:data-stats}, and discuss data preprocessing (\eg, deduplication) in Appendix~\ref{sec:appendix:dataset_preprocessing}. 
Requests span real-world conversation, question answering, translation, code generation, and long-context math reasoning tasks. 

We set up a cluster of 16 NVIDIA A100 GPUs, where requests arrive follow Microsoft's realistic LLM serving trace~\cite{dynamollm-isca24}, scaled by our cluster resource capacity.

% For experiments involving Gemini models, we use the Google Cloud Vertex AI API.
% 
% follow state-of-the-art LLM serving settings~\cite{agrawal2023sarathi, vtc-osdi24} to generate request arrival times using a Poisson distribution.

\begin{figure*}[t]
  \centering
  % First row
  \subfigure[Load Offloading (MS MARCO). \label{fig:offload_msmacro}]{%
    \includegraphics[width=0.24\linewidth]{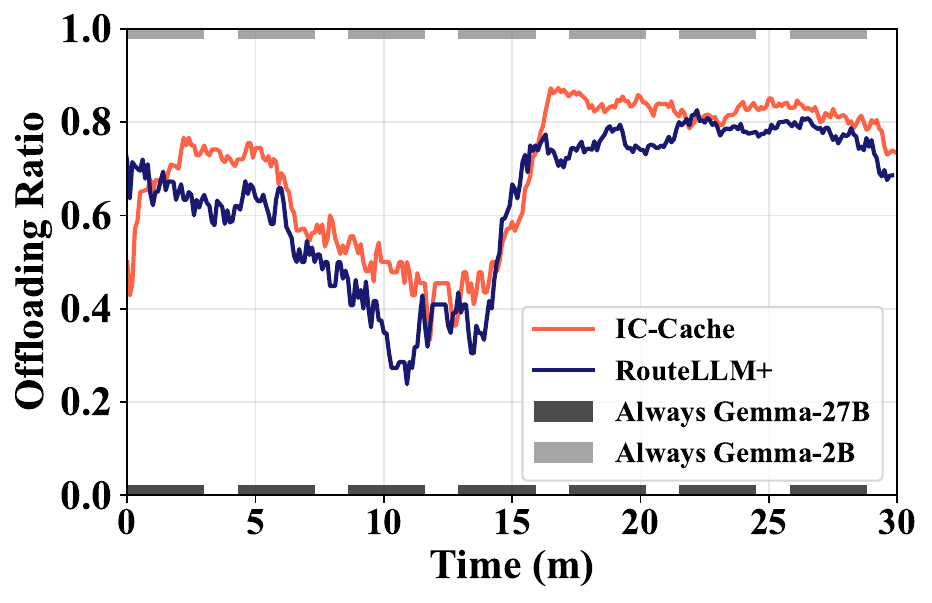}}
  \hfill
  \subfigure[Load Offloading (Natural Questions). \label{fig:offload_nq}]{%
    \includegraphics[width=0.24\linewidth]{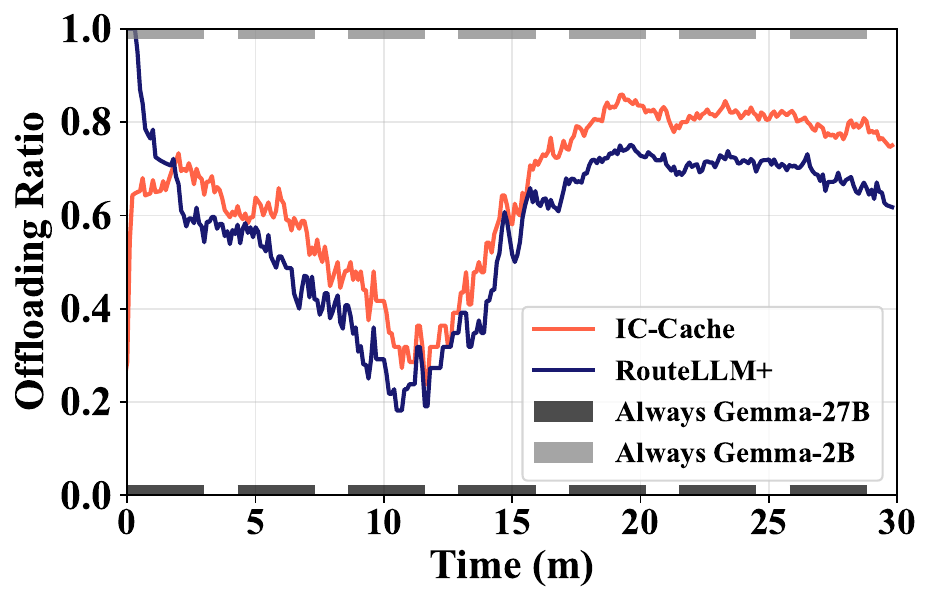}}
  \hfill
  \subfigure[Serving latency (MS MARCO). \label{fig:latency_msmacro}]{%
    \includegraphics[width=0.24\linewidth]{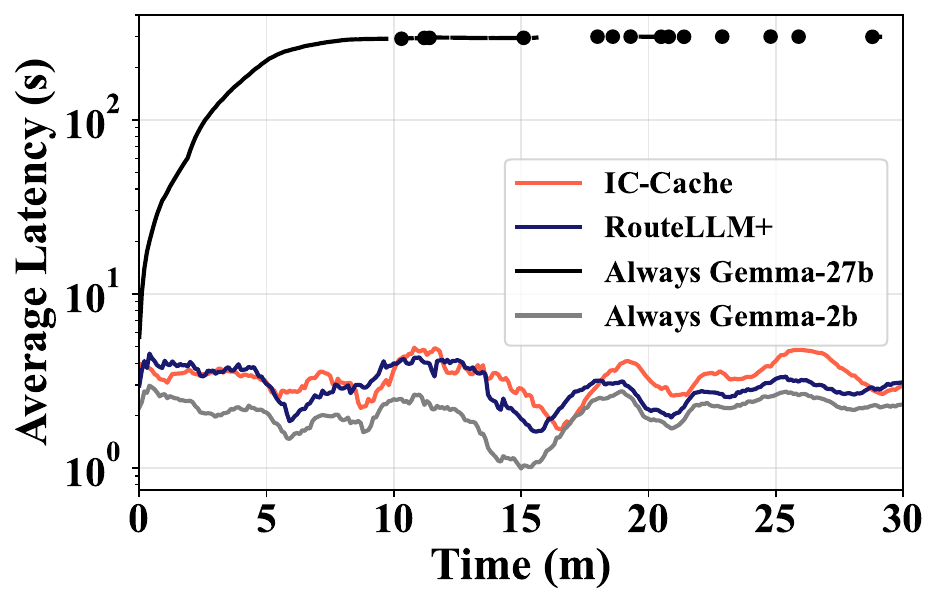}}
  \hfill
  \subfigure[Serving latency (Natural Questions). \label{fig:latency_nq}]{%
    \includegraphics[width=0.24\linewidth]{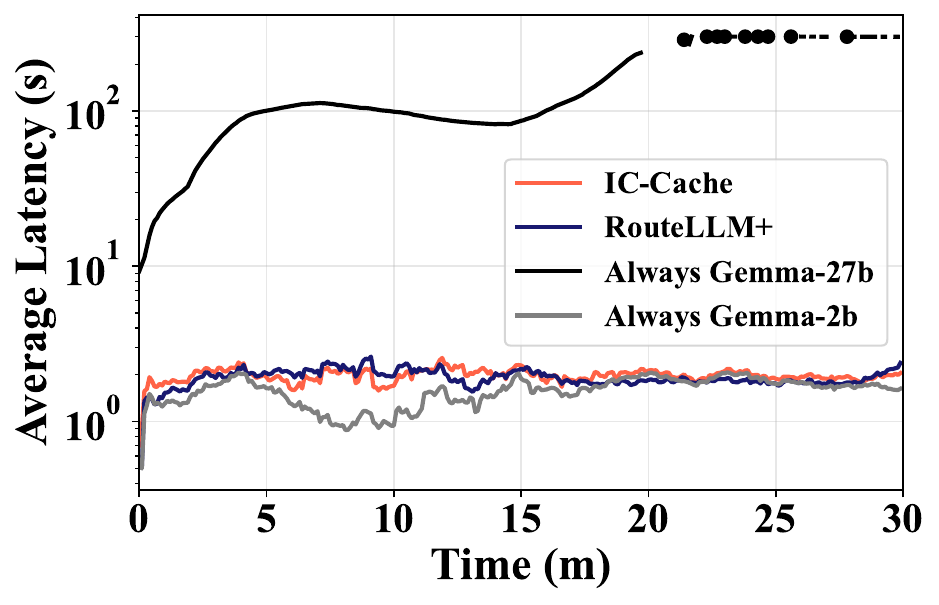}}
  
  % Second row
  \subfigure[Response Quality (MS MARCO). \label{fig:winrate_msmacro}]{%
    \includegraphics[width=0.24\linewidth]{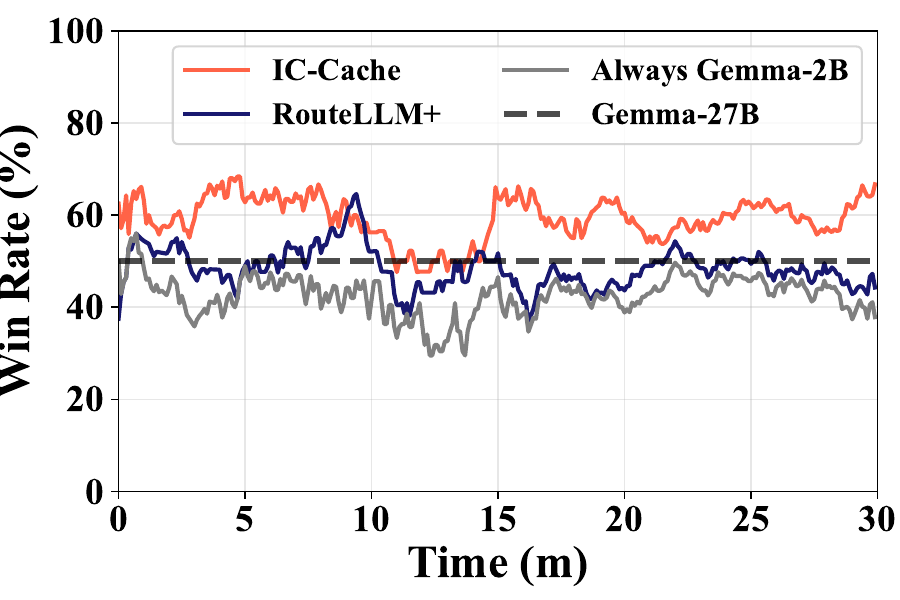}}
  \hfill
  \subfigure[Response Quality (Natural Questions). \label{fig:winrate_nq}]{%
    \includegraphics[width=0.24\linewidth]{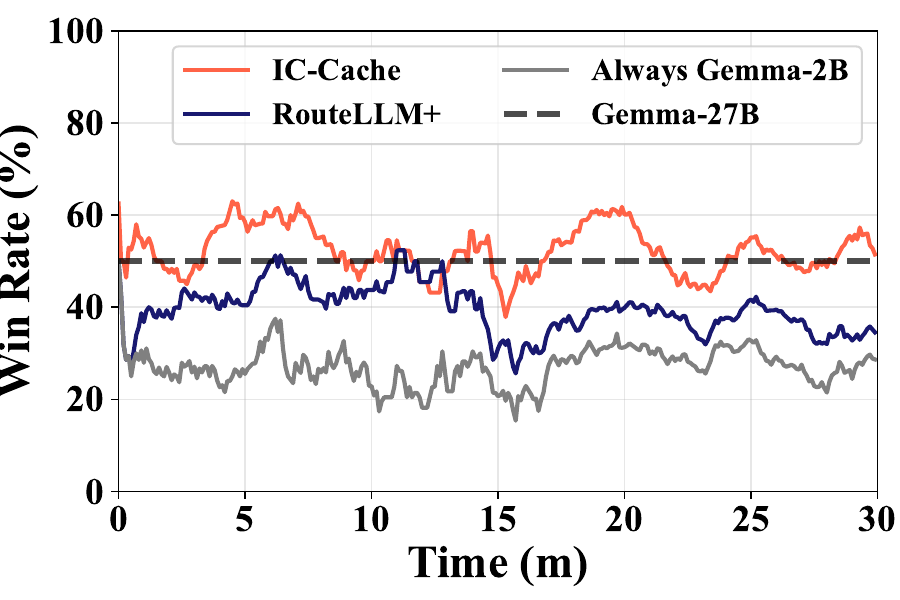}}
  \hfill
  \subfigure[Response Quality (LMSys). \label{fig:winrate_lmsys}]{%
    \includegraphics[width=0.24\linewidth]{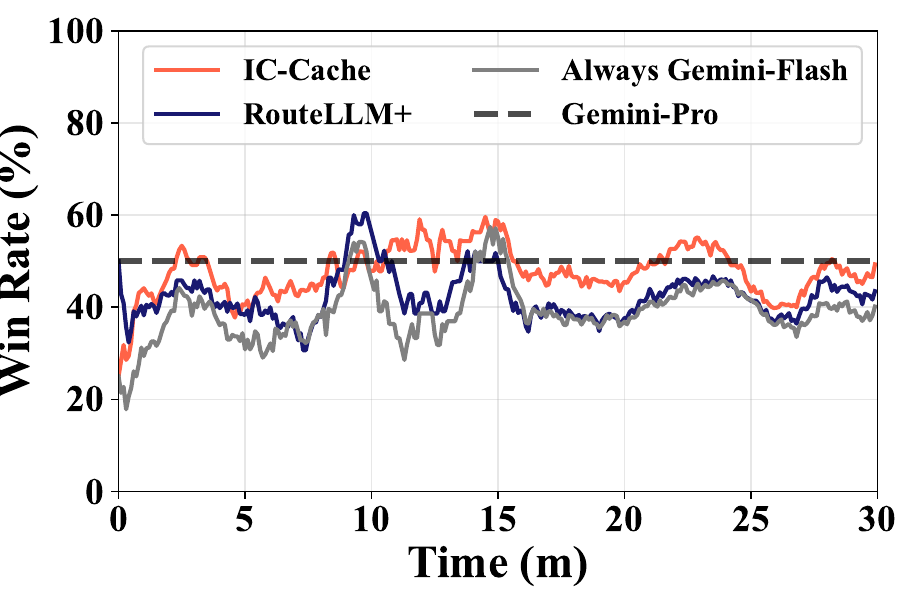}}
  \hfill
  \subfigure[Response Quality (Orca). \label{fig:winrate_orca}]{%
    \includegraphics[width=0.24\linewidth]{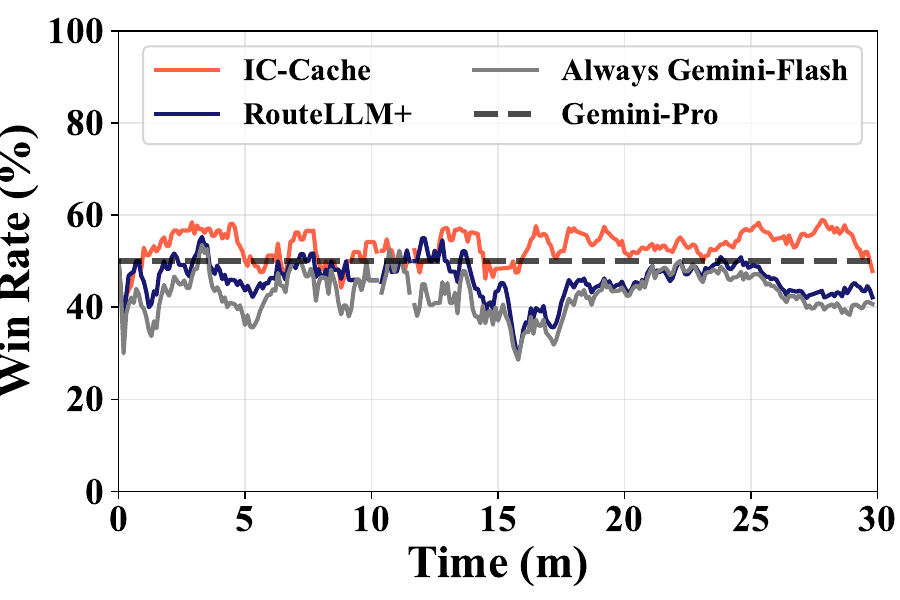}}
  
  \caption{\name enables offloading more requests to small models, improving serving throughput (a)-(b), while achieving better response latency (c)-(d) and response quality (e)-(h).}
  \label{fig:online_e2e_perf}
\end{figure*}

 \begin{table}[t]
\centering
\resizebox{\columnwidth}{!}{  % Makes table fit column width
\begin{tabular}{l|l|c|c}
% \hline
\hline
\textbf{Task} & \textbf{Dataset} & \textbf{Example Size} & \textbf{Request Size} \\
\hline
\multirow{3}{*}{Conversation} &  Alpaca\cite{alpaca} &  32,392 &  1,800 \\
 &  lmsys-chat-1m~\cite{lmsys-chat} &  273,043 & 15,170 \\
 &  OpenOrca~\cite{OpenOrca} & 774,285  & 43,016 \\
\hline

\multirow{2}{*}{Question Q\&A} &  MS MARCO~\cite{DBLP:journals/corr/NguyenRSGTMD16} & 808,731 & 101,092 \\
 &  Natural Questions~\cite{47761} & 300,000  & 7,830 \\
\hline
Translation & WMT-16-PM \cite{bojar-etal-2016-findings}&600,000 & 1000 \\
\hline
Code Gen. & Nl2bash \cite{ye2023generating} & 8090 & 609 \\
\hline
Math reasoning & Math500-Level5\cite{dpsynth}&7500 &5000 \\
\hline
% Math Reasoning & Math \cite{dpsynth}&7500 &5000 \\
% \hline
\end{tabular}
}
\caption{Our evaluation data spans millions of realistic requests. }
\label{tab:data-stats}
\end{table}

\paragraph{Baselines.} 
\label{sec:eval:baselines}
To the best of our knowledge, \name is the first system to optimize LLM serving through real-time capability augmentation using historical requests, complementing existing LLM inference frameworks. We compare \name against the following state-of-the-art baselines:
\begin{itemize} 

\item \emph{w/o \name}: The widely-used LLM serving system (\eg, vLLM~\cite{vllm-sosp23}) without \name integration. 

\item \emph{RouteLLM}~\cite{routellm}: A model routing framework that dynamically selects between a small and a large model based on a binary classifier.  

\item \emph{LongRAG~\cite{longrag-emnlp24}}: Retrieves external documents as auxiliary knowledge. We follow the same retrieval method to select and append the top-5 documents to the prompt. 

\item \emph{Semantic Caching~\cite{databricks-cache}}: Caches past requests and returns cached responses based on embedding similarity.
\end{itemize}

% \name is the first system to support and optimize LLM serving with real-time distillation through cached queries, and is complementary to existing LLM inference frameworks. Our evaluation cover two primary baselines:
% (i) \emph{w/o \name}: llm serving system without \name support, such as vllm.
% (ii) different variants of \name with changes in the router design, retriever design, and example managers. (\S\ref{eval:ablation}). 
% (iii) comparison with retrieval-augmented-generation (RAG). 

% \paragraph{Metrics}
% We will evaluate on the following metrics 

% (i) \emph{win rate improvement} We want to qualitatively compare how \name improves the generation of small models and the overall quality under a given budget. Gemini-1.5-pro will compare each pair of responses generated with \name and without \name and decide whether one response is better than the other or there is an tie. The overall win rate is calculated by the win rate plus half of the tie rate.  

% (ii) \emph{latency improvements} that we can achieve. Specially, we care about the median of TTFT and TBT improvements with \name under a given budget, i.e. how many requests routed to the large model.

% (ii) \emph{throughput improvements} that we can achieve. Specially, we care about the output tokens per seconds and the requests per second with \name under a given budget.

\paragraph{Metrics.} We evaluate \name using  following metrics:

\begin{itemize} 
\item \emph{Quality}: 
To evaluate response quality, we follow established practices~\cite{zheng2023judging, meng2024simpo, llama3-report} to adopt the \emph{LLM-as-a-judge} methodology. Specifically, we use DeepSeek-R1 as the autorater for evaluating Gemini outputs, and Gemini-1.5-Pro for the rest, to minimize self-comparison bias. We report the strong alignment of our autoraters with human preferences in Appendix~\ref{app:llm-judger}. 
Each autorater produces a seven-point score ranging from -3 (significantly worse) to 3 (significantly better), where a score within  [-0.3, 0.3] indicates a tie. To reduce the order bias, we sample eight responses for both input orders and report the average across 16 comparisons. 
We compare overall model quality using \emph{average pairwise scores} and \textit{win rates}=$\frac{(\text{\#wins} + 0.5 \times \text{\#ties})}{\text{\#total}}$. A win rate of 0.5 or an average score of 0 indicates parity between compared models.

% on the evaluated dataset.

% , accounting for the partial advantage of ties in favor of \name.
% Paired response comparisons are performed by generating responses for the same query with and without \name. Each pair is evaluated by the expert model, which categorizes the outcome as a win (better with \name), loss (better without \name), or tie (equally good) and assigns an overall quality score to each response. 

\item \emph{Latency}: The latency reduction in Time-To-First-Token (TTFT) and \update{Time-Between-Tokens (TBT).} 
% Time-To-Backlog-Token (TBT). 
% The analysis also includes the fraction of requests routed to larger models.

\item \emph{Throughput}: We quantify the throughput performance in terms of requests per second achieved. 

\end{itemize}

\subsection{End-to-End Performance}
\label{eval:e2e}

\paragraph{\name adapts effectively online for better efficiency.}
% We first evaluate \name under realistic online load conditions. Request arrivals follow a 24-hour trace from Microsoft's LLM serving traces~\cite{dynamollm-isca24}, where arrival timestamps are mapped to requests from the MS MARCO and Natural Questions datasets. Due to space constraints, we report results for the Gemini and Gemma model families here and additional evaluations are provided in Appendix~\ref{app:eval_details}.

\update{We first evaluate \name under realistic online load conditions. Request arrivals follow a 30-minute trace from Microsoft's LLM serving traces (Figure~\ref{fig:request_arrival_pattern}), where arrival timestamps are mapped to requests from the MS MARCO, Natural Questions, LMSys-Chat, and Open Orca datasets. Due to space constraints, we report results for the Gemini and Gemma model families here, and additional evaluations are provided in Appendix~\ref{app:eval_details}. 
}

% We preserved the trace's 24-hour diurnal pattern but compressed the timeline into a 30-minute experimental window.

% Figure~\ref{fig:online_e2e_perf} presents the online offload ratio, serving latency, and response quality under real-time load. \name improves serving throughput and latency by \todo{X} and \todo{Y}, respectively, while achieving 9.0 percent higher response quality compared to RouteLLM. 

\update{
Figure~\ref{fig:online_e2e_perf} presents the online offload ratio, serving latency, and response quality under real-time load. Notably, \name achieves a 9.0\% higher response quality compared to RouteLLM, while maintaining comparable serving latency and throughput.
Without dynamic offloading, relying solely on the small model leads to subpar responses, whereas using only the large model risks overload and increased latency. While RouteLLM offloads requests based on request difficulty, it is oblivious to the current system load. In contrast, \name dynamically adjusts the offloading ratio based on real-time load and request characteristics. It achieves response quality comparable to always using the large model, without incurring the associated cost and latency.}

\begin{figure}[t]
    \centering
    \includegraphics[width=0.9\linewidth]{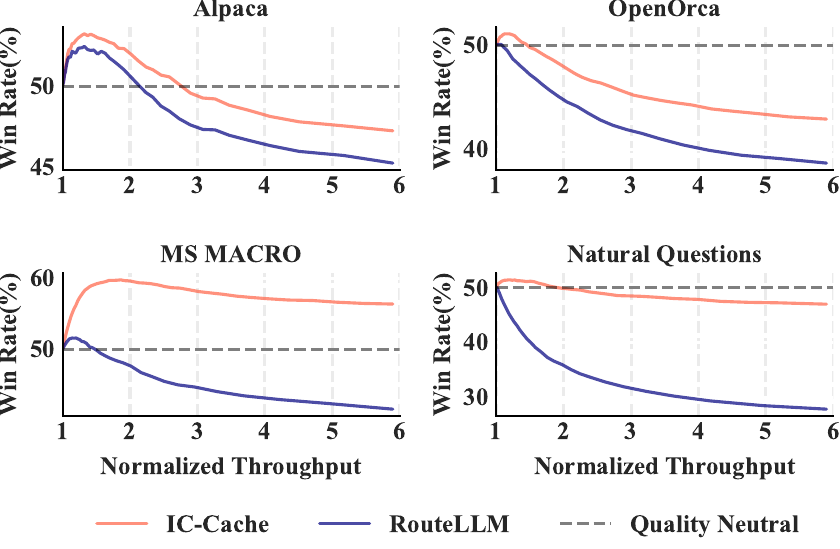}
    \caption{\name enables better quality-efficiency tradeoffs.}
    \vspace{-.2cm}
    \label{fig:quality_throughput_tradeoff}
\end{figure}

\paragraph{\name enables better quality-efficiency tradeoff.}
We further demonstrate that \name enables better quality-efficiency tradeoffs by being aware of the current serving load and intelligently repurposing examples. Figure~\ref{fig:quality_throughput_tradeoff} reports the win rates of Gemma-2-2B over Gemma-2-27B across four datasets, with throughput normalized to that of serving solely with the large model. By varying the router's decision threshold, we dynamically control the fraction of requests offloaded to the small model, investigating the quality and efficiency performance under different requirements on the tradeoff. 

\name consistently achieves higher serving throughput than RouteLLM for the same target response quality. For example, on the Natural Questions dataset, it delivers 2.3$\times$ higher throughput when aiming for a 50\% win rate. Conversely, for a fixed throughput target (e.g., 6$\times$ the throughput of the large model), \name improves response quality by 4--16\%. Notably, by routing each query to the most suitable model, \name enables the 2B model to surpass the 27B model on MS MARCO, achieving a win rate above 50\%.

\begin{figure}[t]
  \centering
    \begin{minipage}{0.48\linewidth}
    \centering
    \includegraphics[width=\linewidth]{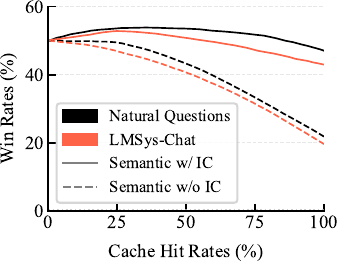}
    \caption{\name augments semantic caching deployment.}
    \label{fig:semantic_ic_eval}
  \end{minipage}
  \hfill
    \begin{minipage}{0.48\linewidth}
    \centering
    \includegraphics[width=\linewidth]{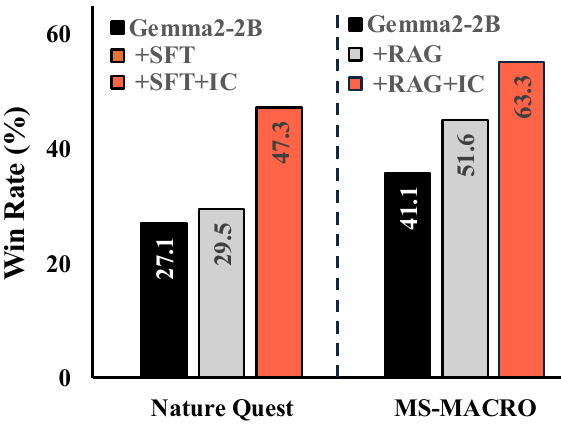}
    \caption{\name augments supervised fine-tuning (SFT) and RAG deployments.}
    \label{fig:sft_rag}
  \end{minipage}
\vspace{-.3cm}
\end{figure}

\paragraph{\name augments existing serving infra effectively.} 
Figure~\ref{fig:semantic_ic_eval} shows that combining \name with existing semantic caching systems consistently improves response quality across different cache hit rates. Higher hit rates, achieved by relaxing the similarity threshold, typically reduce response quality due to less relevant cache matches. In such cases, \name repurposes retrieved responses as in-context examples to improve small LLM outputs, leading to up to a 28\% quality improvement, or equivalently, 4.1$\times$ higher efficiency (hit rate) for the same quality target.

Figure~\ref{fig:sft_rag} further shows that \name augments the popular supervised fine-tuning (SFT) and RAG systems on the Nature Question and MS-MACRO dataset, respectively. 
We notice that while both post-training support, SFT and RAG, \update{improve} the response quality, combining them with \name improves the small model's response quality by 17.8 and 9.8\%, respectively, significantly narrowing the gap between small model performance and that of the larger model.

\subsection{Performance Breakdown}
\label{eval:breakdown}

\begin{figure}[t]
    \centering
    \includegraphics[width=1\linewidth]{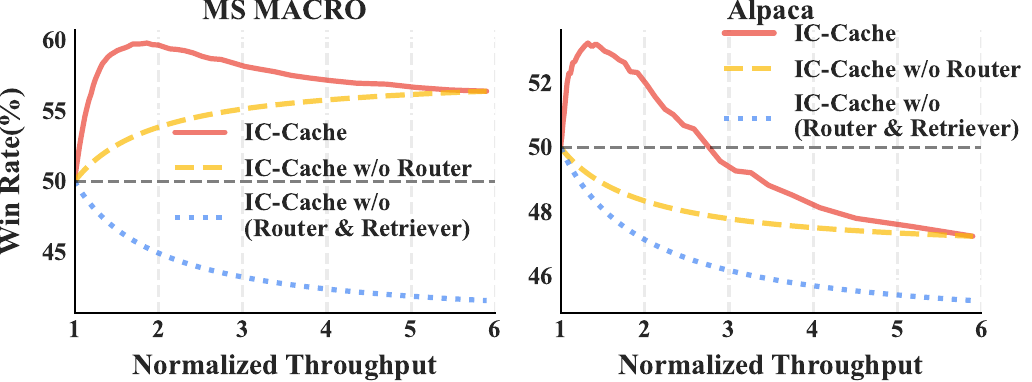}
    \caption{\name achieves better quality-efficiency tradeoff by orchestrating its design components. }
    \label{fig:quality_throughput_tradeoff_breakdown}
\end{figure}

\begin{figure*}[t]
    \centering
    \includegraphics[width=\textwidth]{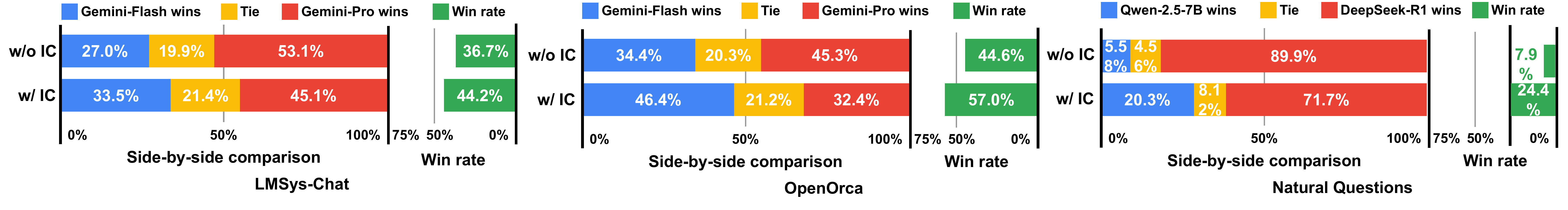}
    \caption{\name improves the quality of generation across different tasks for Gemini, Qwen, and DeepSeek series models.}
    \vspace{-0.1in}
    \label{fig:e2e_quality}
\end{figure*}

\paragraph{Breakdown by components.}
Figure~\ref{fig:quality_throughput_tradeoff_breakdown} shows that \name achieves a superior quality-efficiency tradeoff by effectively orchestrating its design components. Leveraging in-context examples, Gemma-2-2B + \name consistently outperforms Gemma-2-27B on both the MS MARCO and Alpaca datasets. The combination of the request router and example-based augmentation enables \name to establish a new Pareto frontier, surpassing baseline approaches.
Specifically, \name attains a win rate of up to 60\% against Gemma-2-27B while delivering 2$\times$ higher throughput on MS MARCO. On Alpaca, it achieves a 2.8$\times$ throughput improvement with no quality degradation. Importantly, we observe that efforts to increase throughput without the request router lead to suboptimal response quality, highlighting the necessity of quality- and load-aware routing.

% \paragraph{Breakdown by individual response quality.}
We further dissect \name's performance by isolating the effect of repurposing historical examples, \ie, disabling the request router.
Figure~\ref{fig:e2e_quality} reports that \name significantly improves response quality across diverse model families and datasets.
For example, \name improves the win rate of smaller models over larger ones by up to 12.4\% for Gemini models on LMSys-Chat and OpenOrca. Remarkably, on some datasets, smaller models surpass the 50\% win rate, indicating they can outperform their larger counterparts when empowered with high-quality in-context examples. Even in settings with considerable differences in latency, cost, and base accuracy, such as DeepSeek-R1 vs. Qwen2.5-7B, \name still yields an 18\% accuracy gain in win rate.

Beyond the average performance, we observe that \name shifts the entire score distribution toward higher quality, without sacrificing individual requests' quality. Additional results in Appendix~\ref{app:eval_details} confirm that \name provides consistent benefits across a broad range of models and datasets, including Qwen, Gemma, and Phi models.

\begin{figure}[t]
  \centering
    { 
    {\includegraphics[width=0.8\linewidth]{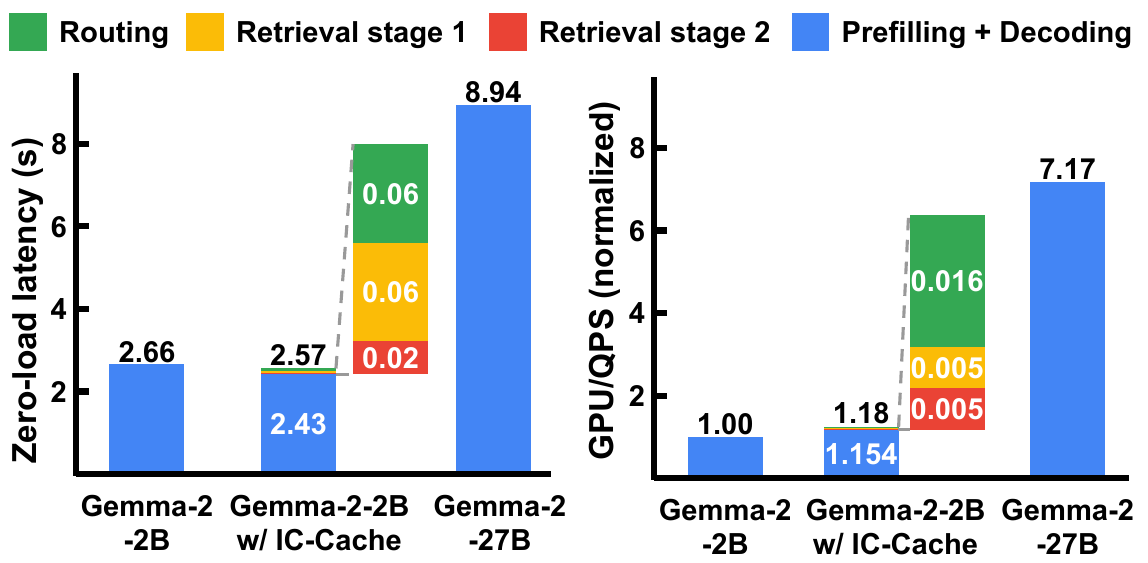}}
    % \hfill
    % \subfigure[Gemini \label{fig:direct_cache_reuse}]{\includegraphics[width=0.48\linewidth]{Figures/eval/moderate_gemini_overhead_breakdown.pdf}} 
    
  }
  \caption{\name introduces little overhead (left) while improving cost-efficiency in sustaining serving throughput (right).}
  \label{fig:perf_breakdown}
\end{figure}

\paragraph{Breakdown by execution lifecycle.}
Figure~\ref{fig:perf_breakdown} presents a detailed execution breakdown for Gemma models. The left figure reports the average contention-free serving latency without batching. We observe that Gemma-2-2B + \name achieves a 3\% reduction in latency compared to Gemma-2-2B alone, attributed to shorter average decoding lengths guided by examples from the large model. Moreover, it is 71\% faster than Gemma-2-27B, primarily due to the smaller model size. 
The right figure illustrates the serving cost, measured as the number of GPUs required to sustain the throughput target, \ie, normalized to the cost of serving with Gemma-2-2B. Under the same resource constraint, Gemma-2-2B + \name delivers a 5.1$\times$ improvement in system throughput compared to Gemma-2-27B, while introducing negligible overhead.

\subsection{Ablation Study}
\label{eval:ablation}

%%%% Temporally comment out

% \begin{figure}[t]
%     \centering
%     \includegraphics[width=0.8\linewidth]{Figures/wildchat_32b.png}
%     \caption{\name shifts traffic to higher throughput models under higher QPS }
%     \label{fig:wildchat_qps_regime}
% \end{figure}

% \paragraph{Impact of Router}
% We evaluated our Request Router's  ability to dynamically allocate requests among multiple LLMs (Qwen 1.5B, 7B, 14B and 32B). The evaluation used English conversations from the Wildchat \cite{zhao2024wildchat} dataset, with response quality assessed by Gemini 2.5 Pro. The bandit policy, trained on a 100k sample subset as per the methodology described earlier, was tested on a distinct 10k sample set under varying QPS. Figure~\ref{fig:wildchat_qps_regime} indicates that at low QPS, the bandit primarily utilizes the 14B model which significantly outperforms the 32B model due to IC-Cache. As QPS increases, demonstrating its load-aware capability, the policy progressively shifts traffic towards models with higher throughput, predominantly routing to the 1.5B model under high load conditions to maintain system responsiveness. We note that the 14B model's throughput was constrained in this setup due to long prefill times from included examples; implementing KV caching, omitted in this evaluation, could improve its performance.

% \paragraph{Impact of Replay Budget}

\begin{figure}[t]
  \centering
  \begin{minipage}[t]{0.48\linewidth}
    \centering
    \includegraphics[width=1\linewidth]{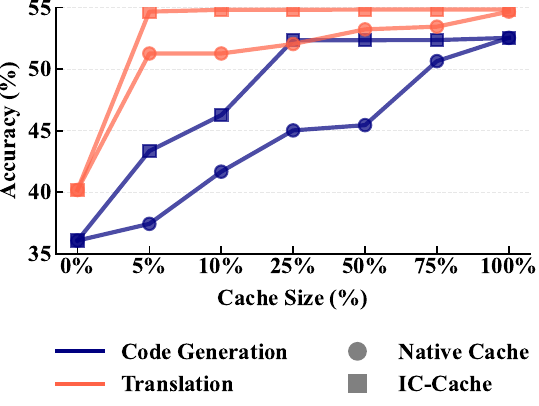}
    \caption{\name delivers improvement under different the example cache sizes.}
    \label{fig:example_pool_breakdown}
    
  \end{minipage}%
  \hfill
    \begin{minipage}[t]{0.48\linewidth}
    \centering
    \includegraphics[width=\linewidth]{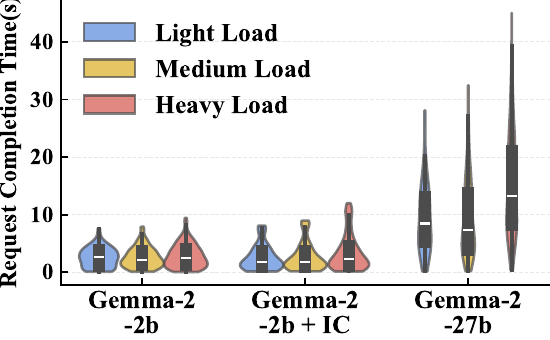}
    \caption{\name improves serving efficiency across serving loads.}
    \label{fig:e2e_latency}
  \end{minipage}
\end{figure}

\paragraph{Impact of example cache size.} 
To evaluate the impact of cache size on generation quality, we conduct experiments using Qwen2.5-3B on both code generation and translation tasks, varying the size of the example cache pool. Specifically, we retain different proportions of the full example set, ranging from 5\% to 100\%, and compare two strategies:
(i) Naive Cache, which randomly retains examples, and
(ii) \name, which employs utility-aware example caching (\S\ref{sec:manager}). 
As shown in Figure~\ref{fig:example_pool_breakdown}, \name achieves near-saturated performance even with a minimal cache, just 12,056 examples for translation and 2,022 for code generation. These correspond to under 20MB in plaintext, demonstrating the efficiency and practicality of our caching design, particularly for memory-constrained deployments.

\begin{figure}[t]
  \centering
  \begin{minipage}[t]{0.48\linewidth}
    \centering
    \includegraphics[width=1\linewidth]{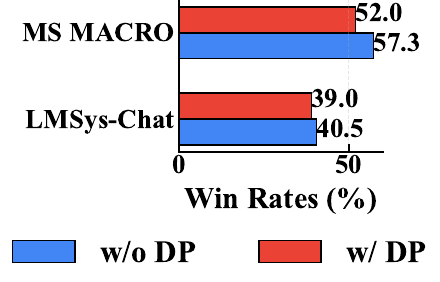}
    \caption{\name with DP synthetic example pool brings marginal quality degradation.
    % on MS MACRO dataset and LMsys-chat dataset of Gemma-2-2B vs. Gemma-2-27B.
    }
    \label{fig:ablation_dp}
  \end{minipage}
  \hfill
  \begin{minipage}[t]{0.48\linewidth}
    \centering
    \includegraphics[width=\linewidth]{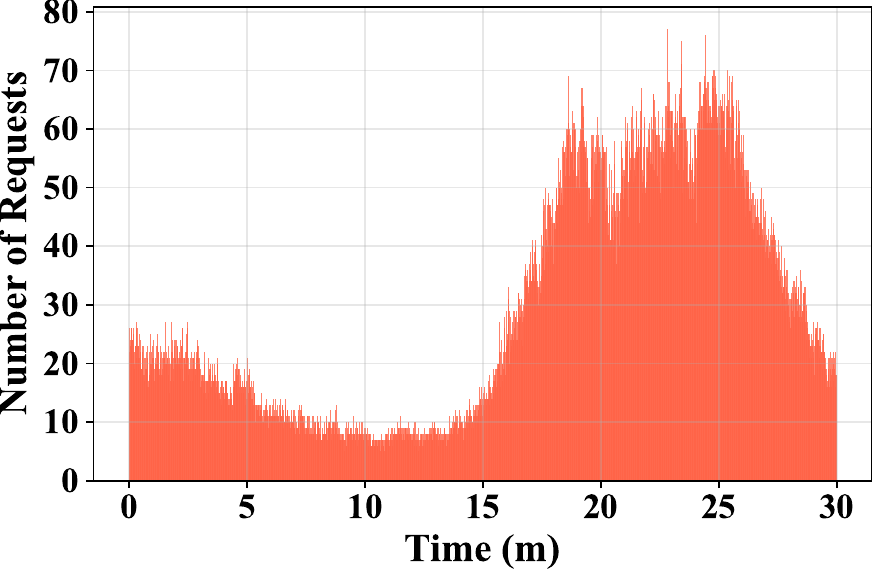}
    \caption{Request arrival pattern sampled from Microsoft's LLM serving traces.}
    \label{fig:request_arrival_pattern}
  \end{minipage}
\end{figure}

\paragraph{Impact of Serving Loads.} 
Figure \ref{fig:e2e_latency} shows that \name achieves superior serving latency performance under varying request loads on the Alpaca dataset.
The light, medium, and high load levels correspond to QPS = 1, 2, and 4, respectively. The end-to-end latency of Gemma-2-2B + \name is similar to that of Gemma-2-2B without \name, with 11--35\% lower P50 latency and 14-31\% higher P99 latency primarily due to different patterns in decoding length distributions, aligning with our observations in Figure~\ref{fig:perf_breakdown}. Compared to Gemma-2-27B, Gemma-2-2B + \name reduces P50 latency by 75--83\% and P99 latency by 69--71\% because of the 10$\times$ difference in model size. We observe similar patterns on other datasets and model families (Appendix~\ref{app:eval_details}). 

% yielded a substantial improvement because of the long-tail access pattern and abundant similarities of examples (\S\ref{sec:manager}), increasing the win rate from 39.29\% to 53.75\%. This result highlights the effectiveness of leveraging even a small number of relevant examples.  Furthermore, we observed a consistent positive trend, with the win rate reaching 56.35\% when utilizing the full cache. So a larger cache size contributes to improved performance, albeit with diminishing returns.

% \begin{figure}[t]
%   \centering
%     { 
%     {\includegraphics[width=0.9\linewidth]{Figures/eval/breakdown/cache_accuracy.pdf}}

%   }
%   \caption{\name delivers improvement under different the example cache sizes. Qwen-3b model on code generation and translation dataset.}
%   \label{fig:example_pool_breakdown}
% \end{figure}

\paragraph{Impact of DP synthesized example pool.}
% \begin{figure}[t]
%   \centering
%     { 
%     {\includegraphics[width=0.8\linewidth]{Figures/eval/ablation/differential_privacy.pdf}}
%     % \hfill
%     % \subfigure[Gemini \label{fig:direct_cache_reuse}]{\includegraphics[width=0.48\linewidth]{Figures/eval/moderate_gemini_overhead_breakdown.pdf}} 
    
%   }  \caption{
%   \name with DP synthetic example pool brings marginal quality degradation on MS MACRO dataset and LMsys-chat dataset of Gemma-2-2B vs. Gemma-2-27B.
%   }
%   \label{fig:ablation_dp}
% \end{figure}
 
To test the effect of using DP synthetic examples instead of the original example pool, we generate DP synthetic examples for MS MACRO and LMsys-chat, two datasets with real user queries where DP synthesis may be considered necessary. 
As shown in Figure~\ref{fig:ablation_dp}, \name's quality slightly decreases with a DP synthetic example pool instead of the original examples, but still improves performance over a non-\name design.

% \begin{table}[t]
% \centering
% \resizebox{\columnwidth}{!}{  % Makes table fit column width
% \begin{tabular}{lcccc}
%         \toprule
%          & Alpaca & LMSys-Chat & MS MACRO & Natural Questions \\
%         \midrule
%         w/o DP & 47.25\%  & 40.51\% & 57.30\% & 47.44\% \\
%         w/ DP & 50.22\% & 38.97\% & 52.03\% & 36.45\% \\
%         \bottomrule
%     \end{tabular}
% }
% \caption{Win rate of Gemma-2-2B + \name over Gemma-2-27B on four datasets using original example pool (w/o DP) and DP synthesized example pool}
% \label{tab:ICC_DP}
% \end{table}

% \begin{table}[t]
% \centering
% \resizebox{0.48\columnwidth}{!}{  % Adjusted width for wider table
% \begin{tabular}{lccc}
%     \toprule
%     Setting & Avg score & Win rates (\%) \\
%     \midrule
%     Gemma-2B & -0.427 & 41.54 \\
%     +RAG & 0.005 & 52.63 \\
%     +RAG+IC & \textbf{0.297} & \textbf{62.40} \\
%     \bottomrule
% \end{tabular}
% }
% \caption{\name augments LongRAG's response quality (MS-MACRO).}
% \label{tab:ICC_RAG}
% \end{table}

\begin{table}[t]
\centering
\resizebox{\columnwidth}{!}{  % Makes table fit column width
\begin{tabular}{lcccc}
        \toprule
         & Gemma-2B & \makecell{Gemma-2B\\+ RAG} & \makecell{Gemma-2B\\+ IC}  & \makecell{Gemma-2B \\ + IC + RAG} \\
        \midrule
        Avg score & -0.4272  & 0.0047 & \underline{0.0667} & \textbf{0.2972} \\
        Win rates(\%) & 41.54 & 52.63 & \underline{56.35} & \textbf{62.40} \\
        \bottomrule
    \end{tabular}
}
\caption{\name complements LongRAG and improves its performance. Gemma-2-2B over Gemma-2-27B on MS MACRO.}
\label{tab:ICC_RAG}
\end{table}

\paragraph{\name vs. RAG.} 
Table \ref{tab:ICC_RAG} shows the average pairwise scores and win rates between Gemma-2-2B and Gemma-2-27B on MS MACRO using a combination of LongRAG (introduced in Section \ref{sec:eval:baselines}) and \name. Both retrieved documents and historical query examples can provide extra information to the model to answer users' queries, which is often more helpful to the smaller models due to their less capacity. \name outperforms RAG because knowledge transfer from historical responses from Gemma-2-27B makes the responses of Gemma-2-2B more aligned with Gemma-2-27B. Such results were also observed in model training~\cite{zhou2024distillspecimprovingspeculativedecoding}. More importantly, \name can be used together with RAG to boost the response quality even further, significantly outperforming RAG-only Gemma-2-27B. 

\paragraph{\name vs. Supervised Finetuning (SFT).}
\label{sec:eval:icc_vs_sft}
To assess the benefits of \name compared to traditional fine-tuning, we finetune Gemma-2-2B on a Natural Questions dataset to mimic the output of the larger Gemma-2-27B model.
Table \ref{tab:ICC_SFT} presents the results on Natural Questions (in-domain task) and Alpaca (out-of-domain task) test sets. While fine-tuning led to some improvements on Natural Questions, the gains were less pronounced than those achieved by \name. 
In contrast, \name's live LLM capability augmentation allows the model to leverage information from the larger model without modifying its own weights, adapting to new domains while preserving its original knowledge.

\begin{table}[t]
% \begin{subtable}
% \centering
% \resizebox{\columnwidth}{!}{  % Makes table fit column width
% \begin{tabular}{lccc}
%         \toprule
%         \makecell{Natural\\Questions} & Gemma-2B & \makecell{Gemma-2B\\+ in-domain SFT} & \makecell{Gemma-2B+\\ in-domain IC} \\
%         \midrule
%         Avg score       & -1.0322   & \underline{-0.8228} & \textbf{-0.2701} \\
%         Win rates(\%)   & 27.87     & \underline{29.11}   & \textbf{47.03}  \\
%         \bottomrule
%     \end{tabular}
% }

% \end{subtable}

\begin{subtable}
\centering
\resizebox{\columnwidth}{!}{  % Makes table fit column width
\begin{tabular}{lcccc}
        \toprule
        Alpaca           & Gemma-2B & \makecell{Gemma-2B\\+ OOD SFT} & \makecell{Gemma-2B\\+ in-domain IC} & \makecell{Gemma-2B\\+ OOD IC}\\
        \midrule
        Avg score       & -0.1896   & -0.5927 & \textbf{-0.1792}  & \underline{-0.2104} \\
        Win rates(\%)   & 45.58     & 32.33   & \textbf{47.25}   & \underline{46.69} \\
        \bottomrule
    \end{tabular}
}
\end{subtable}
\caption{Quality comparison between \name and SFT. Gemma-2-2B vs. Gemma-2-27B on Natural Questions (in-domain) and Alpaca (out-of-domain, OOD).}
\vspace{-.3cm}
\label{tab:ICC_SFT}

\end{table}

%% file: sections/related.tex
\section{Related Work}
\label{sec:related}

\paragraph{LLM Serving Systems.}
Recent LLM serving advances have primarily focused on efficiency. Orca's continuous batching increases throughput~\cite{yu2022orca}, while vLLM \cite{vllm-sosp23} offers LLM execution beyond GPU memory capacity. SARATHI~\cite{agrawal2023sarathi} employs chunked prefill techniques to improve throughput and GPU utilization. FastServe\cite{wu2023fast} proposes a preemptive scheduling to mitigate the queuing delay. Systems like DistServe\cite{zhong2024distserve}, TetriInfer\cite{hu2024inference}, and Splitwise\cite{patel2024splitwise} employ a disaggregation strategy to separate prefill and decode phases for low interference latencies.  PowerInfer-2\cite{xue2024powerinfer} leverages the sparsity of neuron activation to predict and prefetch neurons for on-device LLM serving. 
% Despite these advancements, many systems still face trade-offs between cost and performance, often leading to compromises in either quality or latencies. 
% By exploiting the in-context learning abilities of LLMs and the abundance of past queries, 
\name complements existing LLM serving systems by exploiting the in-context learning abilities of LLMs without altering scheduling order.
% and execution patterns. 
% flexibly enable stable service with superior response quality and less costs according to the budget requirements.

\paragraph{RAG Systems.}
Retrieval-Augmented Generation (RAG) improves the reliability of LLM outputs by integrating knowledge retrieved from external sources\cite{lewis2020retrieval}. 
% RAG encompasses three main processes: retrieval, generation, and augmentation\cite{gao2023retrieval}. In the retrieval stage, RAG systems 
It identifies relevant text chunks using either sparse retrieval methods, such as BM25 \cite{robertson2009probabilistic} and TF-IDF \cite{sparck1972statistical}, or dense retrieval methods. 
% , including Dense Passage Retriever (DPR) \cite{karpukhin2020dense}, Contriever \cite{izacard2021unsupervised}, and Spider \cite{ram2021learning}. 
% RAG can generate outputs with parameter-accessible models, like the Gemma-2\cite{team2024gemma} series, or parameter-inaccessible models, such as ChatGPT-4\cite{achiam2023gpt} and Gemini-1.5-pro\cite{team2023gemini}. 
The retrieval process can be further optimized with techniques like iterative\cite{shao2023enhancing}, recursive\cite{kim2023tree}, or adaptive retrieval\cite{jiang2023active,asai2023self}.
% 
% In addition, generation performance can be improved by fine-tuning generation models\cite{zhang2024raft}.
CacheBlend\cite{yao2024cacheblend} reduces RAG system latency by storing and reusing KV caches with selective recomputations. 
However, RAG relies on long external sources and is vulnerable to out-of-domain or low-quality documents~\cite{leemann2024auto}. 
\name complements RAG by generating cached queries with RAG to incorporate external knowledge (\S\ref{eval:ablation}). 
% Moreover, \name addresses quality-efficiency tradeoffs in flight. 
% with request routing and example caching optimizations. 
% and strengthening output utility through learning from context-aligned examples generated by large models.

\paragraph{Knowledge Distillation and In-context Learning.}
In-context learning (ICL) allows LLMs to perform new tasks by learning from demonstrations in the input context~\cite{brown2020language,wei2022emergent}. The effectiveness of ICL is influenced by multiple factors, including the number of demonstrations, quality, diversity, and order~\cite{luo2024context,brown2020language}. 
Ceil~\cite{ceil-icml23} trains an example selector to pick examples from external documents. 
% While LLMs generally benefit from an increased number of demonstrations, there's a diminishing return for the number of examples\cite{brown2020language, min2022rethinking}. 
% We also show this in the Figure. While increasing the number of demonstrations can enhance LLM performance, the quality and diversity of selected examples are equally vital to achieving optimal results\cite{ye2023compositional}. Additionally, research has shown that demonstration order can significantly impact model performance, with some tasks displaying performance variations ranging from near-random to state-of-the-art depending on the sequence of examples\cite{lu2021fantastically}.  
\name exploits the ICL capability and the high volume of requests in LLM serving systems to optimize the generation quality-efficiency tradeoff with example selection, request routing, and management.

%% file: sections/discussions.tex
\section{Discussions}
\label{sec:discussions}

% \subsection{Handling Model and Data Dynamics}
% One of the key challenges in real-world LLM serving is to adapt to dynamic conditions, such as query distribution shifts and occasional model updates. \name is designed to address those challenges with dynamic example management and model routing, ensuring the long-term effectiveness and efficiency. 

\paragraph{Handling Query Distribution Shift. }\update{User interests and popular topics are not static. They can cause the query distribution to shift over time. \name is designed for this with its two core components. First, the MAB-based request router is inherently adaptive to distribution shift. Unlike a static classifier that would degrade as the query distribution drifts from its training data, the bandit model continuously learns from recent requests. It adjusts its routing policy in a data-efficient manner, adapting to evolving topics and changing example utility without requiring costly offline retraining. Second, the Example Manager actively refines caches and evicts stale queries to reflect current query trends. Its eviction policy is aware of the cost and time to live of each query and response pair, which ensures that stale or low-utility examples are replaced with fresh, relevant ones. By maintaining a moving average of each example's utility with a decay factor, the system prioritizes examples that are effective for the current query distribution, ensuring the cache's contents do not become obsolete.}

\paragraph{Handling Model Updates.} \update{LLM providers can update their models from time to time, which can alter performance characteristics and make static routing policies suboptimal. \name's architecture provides inherent resilience to such changes. The Request Router's exploration-exploitation strategy allows it to dynamically probe the performance of updated models and adjust the traffic accordingly. For instance, if a smaller model is upgraded and can suddenly handle a new class of queries effectively with augmentation, the router will detect this shift through online feedback and begin offloading more requests to it. This avoids the prohibitive cost of generating and re-labeling vast datasets each time a model is updated, ensuring the system can fluidly accommodate improvements or changes in the underlying LLM fleet.}

\paragraph{Performance and Quality Tradeoff.} \update{\name relies on online capability augmentation, which is grounded in in-context learning theory. It allows for on-the-fly knowledge transfer and imitation without costly retraining the model. By providing high-quality demonstrations, \name guides the smaller model to generate responses that are not only structurally similar but also capture the important information of the larger model, thus maintaining high quality while benefiting from the smaller model's lower latency and resource footprint. When multiple models are available, we can identify more sweet spots on the efficiency-quality curve with offline profiling. Instead of being limited to a binary choice between a single small, fast model and a single large, high-quality one, the request router can select the most appropriate model.
% for a given task's complexity.  
}

%% file: sections/conclusion.tex
\section{Conclusion}
\label{sec:conclusion}

We introduce \name, an in-context caching system for LLM serving that leverages historical requests as in-context examples. \name identifies high-utility examples and efficiently prepends them to the input for better response at scale. \name employs a cost-aware cache replay mechanism to improve example quality offline, and a bandit-based request router to adaptively route requests to LLMs with varying capabilities. Our evaluations on real-world datasets demonstrate that \name improves both serving throughput and latency.
% while maintaining response quality.

%% file: sections/appendix.tex
% At the location where you want the table
\section{Appendix}
% \appendix
\subsection{Prompt Templates}
We list all system prompt templates for the generative models with or without IC-Cache, and the autoraters in Figure \ref{fig:prompt_no_ic_cache} -- \ref{fig:prompt_autorater}. 

\begin{figure*}[h!]
\begin{llmprompt}[label={text:ic-cache-prompt}]{System Prompt without IC-Cache} % Example with optional label
[System]

You are a helpful AI Assistant that follows users' instructions carefully. Write a response that appropriately completes the request. Provide necessary details or explanations if that helps to exceed the user's expectations. 

\vspace{\baselineskip}

Below is an instruction that describes a task: 

\{instruction\}

\end{llmprompt}
\caption{System Prompt without IC-Cache for conversational tasks.}
\label{fig:prompt_no_ic_cache}
\end{figure*}

\begin{figure*}[h!]
\begin{llmprompt}[label={text:ic-cache-prompt}]{System Prompt with IC-Cache} % Example with optional label
[System]

You are a helpful AI Assistant that follows users' instructions carefully. Write a response that appropriately completes the request. Provide necessary details or explanations if that helps to exceed the user's expectations. 

\vspace{\baselineskip}

Below is an instruction that describes a task: 

\{instruction\}

\vspace{\baselineskip}

Below are examples of detailed instructions and responses. When a user gives you an instruction, consider the following:

**Relevance: Do the examples directly relate to the user's specific task or question? If not, focus on completing the user's request without relying on the examples.

**Quality: Do the examples demonstrate excellent explanations, detail, and clarity? If so, you may follow their format and style to improve your own response.

**Helpfulness: Do the examples provide helpful information that is relevant to the user's instruction? If so, you may use the information in the examples to help you complete the user's instruction.

\vspace{\baselineskip}

\{examples\}

\vspace{\baselineskip}

Below is an instruction that describes a task. 
Write a response that appropriately completes the request. Provide necessary details or explanations if that helps to exceed the user's expectation. Remember: Your primary goal is to understand the user's instruction and complete the task with informative detail. The examples are resources to guide you, not strict templates to follow. However, you can refer to and follow the examples if the user's instruction is very similar to the examples. 

\vspace{\baselineskip}

Below is an instruction that describes a task again:

\{instruction\}
\end{llmprompt}
\caption{System Prompt with IC-Cache for conversational tasks.}
\label{fig:prompt_ic_cache}
\end{figure*}

\begin{figure*}[h!]
\begin{llmprompt}[label={text:autorater-prompt}]{Autorater System Prompt} %
[System]

Please act as an impartial judge and evaluate the overall quality of the responses provided by two AI assistants to the user question displayed below. You should choose the assistant that follows the user's instructions and answers the user's question better. Your evaluation should consider factors such as instruction following, factuality, helpfulness, depth, creativity, and level of necessary details of their responses. Avoid any position biases and ensure that the order in which the responses were presented does not influence your decision. Do not allow the length of the responses to influence your evaluation. Do not favor certain names of the assistants. Be as objective as possible. 

\vspace{\baselineskip}

You should start with your evaluation by comparing the two responses and provide a short rationale. After providing your rationale, you should output the final verdict by strictly following this seven-point Likert scale: 3 if assistant A is much better, 2 if assistant A is better, 1 if assistant A is slightly better, 0 if the two responses have roughly the same quality, -1 if assistant B is slightly better, -2 if assistant B is better, and -3 if assistant B is much better.

\vspace{\baselineskip}

You should format as follows:

[Rationale]: Placeholder for the short rationale of the score. (less than 200 words)

[Score]: Placeholder for the score. This should be -3, -2, -1, 0, 1, 2, or 3.
\end{llmprompt}
\caption{Autorater system prompt for side-by-side quality evaluation.}
\label{fig:prompt_autorater}
\end{figure*}

\subsection{Sample complexity of router training}
\label{app:sample-complexity}

As indicated in the design section, we always pick the highest confidence score LLM and select the second LLM via Thompson sampling. If the sample has been selected for training, that implies ambiguity, and we train the critic model on the response outcomes of the picked models.

Thompson sampling maintains a Beta distribution for each model, representing our belief about its performance. After each comparison or round, we update these distributions and sample from them to make selections.

We assume the Bradley-Terry model for our analysis.
 
\subsubsection{Definitions}
\begin{itemize}
    \item Let $U_i$ be the true utility (quality) of model $i$.
    \item Let $C_i$ be the computational cost (e.g., latency) of model $i$.
    \item Let $L$ be the current system load (e.g., queries per second).
    \item Let $\mu_i(t)$ be the estimated utility of model $i$ after $t$ rounds.
    \item Let $\Delta_i = U_1 - U_i$ where model 1 is the best model in terms of utility.
\end{itemize}

\subsubsection{Convergence Guarantees}

\textbf{Theorem 1:} With the hybrid Thompson sampling approach, the probability of failing to identify the best model after $T$ rounds decreases with $T$ as follows:
\begin{equation}
P(\hat{i}_T \neq 1) \leq (N-1)T^{-C}
\end{equation}
where $\hat{i}_T$ is the model with the highest estimated utility after $T$ rounds, $N$ is the number of models, and $C$ is a positive constant.

\textbf{Proof:}
The proof proceeds by bounding the probability of mistaking a single suboptimal model for the best one, and then summing these probabilities using a union bound.

\begin{enumerate}
    \item \textbf{Union Bound:} The overall probability of failure is the probability that any suboptimal model $i$'s estimated utility $\mu_i$ is greater than the best model's estimated utility $\mu_1$. We can bound this with the union bound:
    \begin{equation}
    P(\hat{i}_T \neq 1) \leq \sum_{i=2}^{N} P(\mu_i > \mu_1)
    \end{equation}

    \item \textbf{Number of Samples:} For Thompson sampling, each suboptimal model $i$ is sampled approximately $O(\log(T)/\Delta_i^2)$ times in $T$ rounds \cite{pmlr-v23-agrawal12}. We can state this more formally for the number of comparisons, $m_i(T)$, for a sufficiently large T:
    \begin{equation}
    m_i(T) \geq K \frac{\log(T)}{\Delta_i^2}
    \end{equation}
    where $K$ is a positive constant.

    \item \textbf{Applying Hoeffding's Inequality:} Let's analyze the probability of a single error, $P(\mu_i > \mu_1)$. Let the empirical difference be $\hat{\Delta}_i(m) = \mu_1 - \mu_i$ after $m$ comparisons, whose true mean is the utility gap $\Delta_i = U_1 - U_i$. The error event $\mu_i > \mu_1$ is equivalent to $\hat{\Delta}_i(m) < 0$. This can be written as a deviation from the true mean: $\hat{\Delta}_i(m) - \Delta_i < -\Delta_i$. Applying Hoeffding's inequality with $\epsilon = \Delta_i$:
    \begin{equation}
    P(\mu_i > \mu_1) = P(\hat{\Delta}_i(m) < 0) \leq e^{-2m_i(T)\Delta_i^2}
    \end{equation}
    
    \item Now we substitute the bound for the number of samples $m_i(T)$:
    \begin{equation}
    P(\mu_i > \mu_1) \leq \exp\left(-2 \left(K \frac{\log(T)}{\Delta_i^2}\right) \Delta_i^2\right)
    \end{equation}
    Which simplifies to:
    \begin{equation}
    P(\mu_i > \mu_1) \leq e^{-2K\log(T)} = e^{\log(T^{-2K})} = T^{-2K}
    \end{equation}
    
    \item Setting $C = 2K$ ,we have $P(\mu_i > \mu_1) \leq T^{-C}$. Substituting this result back into the union bound from step 1 gives the final bound.
\end{enumerate}

\vspace{\baselineskip}

\textbf{Theorem 2:} To identify the best model with probability at least $1-\delta$, the hybrid Thompson sampling approach requires:
\begin{equation}
T = O\left(\frac{N}{\Delta_{\min}^2} \log\left(\frac{N}{\delta}\right)\right)
\end{equation}
comparisons, where $\Delta_{\min} = \min_{i>1} \Delta_i$ is the minimum utility gap.

\textbf{Proof:}
To achieve a total failure probability of at most $\delta$, we use a union bound to require the failure probability for each of the $N-1$ suboptimal models to be at most $\delta/(N-1)$. For a single model $i$, we need to find the number of comparisons $m$ such that $P(\mu_i > \mu_1) \le \delta/(N-1)$. From Hoeffding's inequality, this implies $e^{-m\Delta_i^2/2} \le \delta/(N-1)$. Solving for $m$ using the worst-case gap $\Delta_{\min}$ gives $m = O(\frac{1}{\Delta_{\min}^2}\log(\frac{N}{\delta}))$. The total number of samples $T$ is $(N-1) \cdot m$, which gives the final bound.

\vspace{\baselineskip}

\textbf{Theorem 3:} To identify the top-k models with probability at least $1-\delta$, the hybrid Thompson sampling approach requires:
\begin{equation}
T = O\left(\frac{N}{\Delta_{\min,k}^2} \log\left(\frac{k(N-k)}{\delta}\right)\right)
\end{equation}
comparisons, where $\Delta_{\min,k} = \min_{i \leq k, j > k} (U_i - U_j)$ is the minimum gap between any model in the top-k and any model outside the top-k.

\textbf{Proof:}
A failure occurs if any of the $k(N-k)$ critical pairs (one model from the top-k, one from outside) are incorrectly ordered. Using a union bound, the required error rate for any single pair is $\delta/(k(N-k))$. The critical gap for this problem is $\Delta_{\min,k}$. Applying Theorem 2 with the critical gap yields the stated complexity.

\vspace{\baselineskip}

\textbf{Theorem 4:} Let the score (logit) for selecting model $i$ be defined by its utility and a load-dependent cost penalty:
\begin{equation}
S_i(L) = \mu_i - \lambda_0 \tanh(\gamma L) C_i
\end{equation}
where $\lambda_0, \gamma > 0$ are constants. Let the selection policy be a softmax over these scores. If $j$ is the model with the highest load-adjusted utility as defined, then $\lim_{L \to \infty} P_j(L) = 1$.

\textbf{Proof:}
The selection probability for model $i$ is $P_i(L) = \frac{\exp(S_i(L))}{\sum_{n=1}^N \exp(S_n(L))}$.
\begin{enumerate}
    \item \textbf{Limiting Behavior of Tanh:} As the load $L \to \infty$, the hyperbolic tangent term approaches its maximum value: $\lim_{L \to \infty} \tanh(\gamma L) = 1$.
    \item \textbf{Ratio of Probabilities:} Consider the ratio of probabilities between any model $k \neq j$ (where $C_k > C_j$) and the minimum-cost model $j$:
    \begin{equation}
        \frac{P_k(L)}{P_j(L)} = \frac{\exp(S_k(L))}{\exp(S_j(L))} = \exp(S_k(L) - S_j(L))
    \end{equation}
    Substituting the score definition:
    \begin{equation}
        S_k(L) - S_j(L) = (\mu_k - \mu_j) - \lambda_0 \tanh(\gamma L) (C_k - C_j)
    \end{equation}
    \item \textbf{Asymptotic Limit:} We take the limit of this difference as $L \to \infty$:
    \begin{equation}
        \lim_{L \to \infty} (S_k(L) - S_j(L)) = (\mu_k - \mu_j) - \lambda_0 (C_k - C_j)
    \end{equation}
    Since $C_k > C_j$, the term $(C_k - C_j)$ is a positive constant. For a sufficiently large cost-sensitivity parameter $\lambda_0$, the negative cost term $-\lambda_0(C_k - C_j)$ will dominate the constant utility difference $(\mu_k - \mu_j)$, making the entire expression approach $-\infty$.
    \item As the exponent approaches $-\infty$, the ratio of probabilities approaches zero:
    \begin{equation}
    \lim_{L \to \infty} \frac{P_k(L)}{P_j(L)} = 0 \quad \text{for all } k \neq j
    \end{equation}
    Since $\sum_{i=1}^N P_i(L) = P_j(L) + \sum_{k \neq j} P_k(L) = 1$, and the ratio of every other model's probability to model $j$'s probability goes to zero, it must be that $\lim_{L \to \infty} P_j(L) = 1$.
\end{enumerate}  

\subsection{An Example of IC-Cache}
Figure \ref{fig:ic-cache-example} shows an example of how IC-Cache improves the response quality of the small model with retrieved examples from the large model, collected from the Natural Questions dataset. Without in-context examples, Gemma-2-2B cannot correctly answer the user's question because it has less capacity to memorize certain historical events compared to the larger model. In-context examples similar to the question, not being semantically identical though, provide related information from different perspectives to answer the question even better than the large model. It is also noteworthy that a well-tuned small model like Gemma-2-2B has the capability to distinguish nuanced concepts and avoid distractions even if the retrieval process doesn't have perfect precision (e.g., the unrelated examples in red in Figure \ref{fig:ic-cache-example}). 

\begin{figure*}

\begin{llmprompt}{User Query}
\text{\_\_\_\_\_\_\_} led many viking explorations throughout the north atlantic.
\end{llmprompt}

\begin{llmredresponse}{Gemma-2-2B Response}
Thor led many Viking explorations throughout the North Atlantic.
\end{llmredresponse}

\begin{llmgreenresponse}{Gemma-2-27B Response}
\textbf{Leif Erikson} led many Viking explorations throughout the North Atlantic. He is particularly famous for reaching North America around 500 years before Christopher Columbus. 
\end{llmgreenresponse}

\begin{llmexamples}{In-context Examples (only show queries)}
\begin{itemize}
    \item the viking who explored and settled vinland was \text{\_\_\_\_\_\_\_}
    \item what places did vikings sail to in the north atlantic 
    \textcolor{red}{\item vitus bering what did he explore}
    \textcolor{red}{\item what people of the viking}
    \item what year did leif ericson across the atlantic
\end{itemize}
\end{llmexamples}

\begin{llmgreenresponse}{Gemma-2-2B Response w/ IC-Cache}
\textbf{Leif Erikson} was a Viking explorer who led many expeditions throughout the North Atlantic. He is believed to have reached North America around 1000 AD, establishing the first known Viking settlement in L'Anse aux Meadows in modern Newfoundland, Canada.

\vspace{\baselineskip}

While the exact details of his travels and his motivations are still debated, his voyage marked the beginning of Norse exploration and settlement in what is now Canada and the United States. 

\end{llmgreenresponse}
\caption{An example of how IC-Cache works to improve response quality of Gemma-2-2B using a retrieved example from Gemma-2-27B.}
\label{fig:ic-cache-example}
\end{figure*}

\subsection{Dataset Preprocessing}
\label{sec:appendix:dataset_preprocessing}
The datasets are curated and processed as follows: 
\begin{itemize} 

\item \emph{Preprocessing}: We deduplicate examples and filter out non-English queries, as not all models are multilingual.

\item \emph{Dataset split}: For datasets with predefined training and test splits, we use the training split to populate the example bank and the test split for online request evaluation. Otherwise, we randomly partition the data to kick-start example banks and online request sets. 

\item \emph{Example pool initialization}: For the purpose of experiments within each model family, we initialize the example pool in each dataset using the responses generated by the larger model.
\end{itemize}

% Alpaca, lmsys-chat-1m, and OpenOrca, are conversational datasets to access the models' capability of following instructions, understanding context, and generating human-like coherent responses. MS MACRO and Natural Questions are question answering datasets to evaluate models' breadth of knowledge and ability to comprehend information and provide accurate and relevant answers. 
% Among these, lmsys-chat-1m and MS MARCO contains real user queries submitted to Chatbot Arena~\cite{chiang2024chatbotarenaopenplatform} and Bing Search. \todo{cite datasets here and also in the table}

% Among these, Lmsys-chat and MsMacro datasets contain real user-submitted questions, ensuring evaluation on authentic queries. These three datasets covers different domains, allowing a comprehensive analysis of \name on various use cases: Lmsys-chat collects the queries submitted to the chatbot arena, Natural Questions collects queries submitted to the Google search, and MsMacro dataset collects queries submitted to the Bing Search. In contrast, Alpaca and Orca are synthetic datasets that measure instruction-following abilities. While Alpaca contains only 30k conversations and Orca contains over 300k conversations, we explore how \name performs with varying levels of query abundance. 

\subsection{LLM-as-a-Judge}
\label{app:llm-judger}

\begin{table*}[h] % The ! makes it override most constraints
\small
    \centering
    \begin{tabular}{l | c c c c}
    \hline\hline
    Judge & Gemini-1.5-Flash & Gemini-1.5-Pro & Gemini-2.5-Pro & Human\\
    \hline
    GPT-4  & 74\% & 77\% & 76\% & 66\% \\

    Gemini-1.5-Flash &  & 80\% & 76\% & 67\% \\

    Gemini-1.5-Pro & & & 81\% & 68\% \\

    Gemini-2.5-Pro & & & & 73\% \\

    Human &  &  &  &  63\% \\
    \hline\hline
    \end{tabular}
    \caption{Preference agreement matrix between different LLM raters and human raters on MT-Bench~\cite{zheng2023judging}.}
    \label{tab:mt_bench_agreement}
    
\end{table*}

We adopt an LLM-as-a-Judge policy in our evaluations. To validate the effectiveness of this evaluation method, we assess the alignment between LLM-based judgments and human labels. Specifically, we evaluate the LLM judges used in our study---Gemini-1.5-Pro and Gemini-2.5-Pro---on MT-Bench~\cite{zheng2023judging}, a benchmark designed to measure agreement between model and human preferences. As shown in Table~\ref{tab:mt_bench_agreement}, both Gemini judges exhibit strong alignment with human labels, even outperforming alignments among different human raters. Also, Gemini-1.5-Pro achieves agreement levels comparable to GPT-4, further supporting the reliability of our LLM-as-a-Judge policy.

\section{More Evaluation Results}
\label{app:eval_details}

\begin{figure*}[htb!]
    \centering
    \includegraphics[width=.95\linewidth]{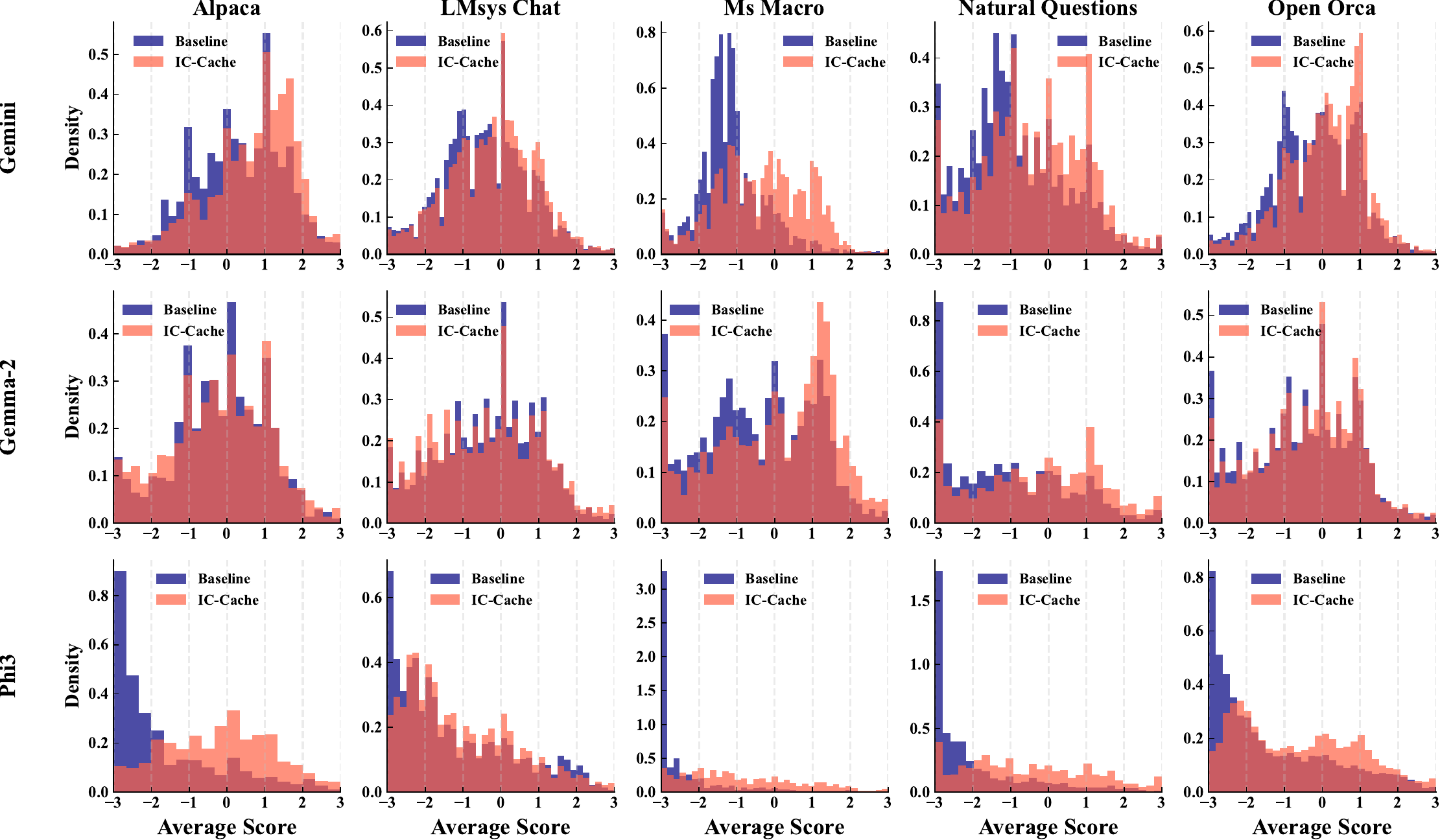}
    \caption{Model quality comparison on five text generation tasks with three different model families. With IC-Cache, the quality of a smaller model can be significantly boosted.}
    \label{fig:quality_improvements_raw_score}
\end{figure*}

% \begin{figure*}[htb]
%     \centering
%     \includegraphics[width=\textwidth]{Figures/eval/e2e/win rates (4).pdf}
%     \caption{\name improves the quality of generation across different tasks for Gemini, Gemma, and Phi-3 series models.}
%     \vspace{-0.1in}
%     \label{fig:e2e_quality_app}
% \end{figure*}

\begin{figure}[t]
  \centering
  \begin{minipage}{0.8\linewidth}
    \centering
    \includegraphics[width=.7\linewidth]{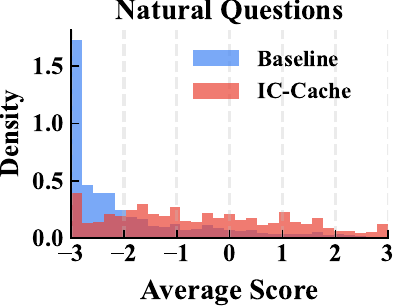}
    \caption{\name improves response score on natural question using Phi-3.}
    \label{fig:phi3_score}
  \end{minipage}
  \hfill
\end{figure}

\subsection{Response Quality Improvement}
\label{sec:appendix:quality_improvements}

IC-Cache brings significant improvements in response quality across diverse model families and datasets (Figure \ref{fig:e2e_quality}). 
% We use side-by-side evaluations with Gemini-1.5-Pro to assess the relative quality of responses between small and large models, with and without IC-Cache.
In this experiment, the model router was configured to route each query to both the small and large models, enabling a direct quality comparison. With IC-Cache, the win rate of smaller models over larger models improves by up to 12.4 percentage points for Gemini models on LMSys-Chat and OpenOrca. Notably, on certain datasets, IC-Cache enables smaller models to achieve win rates exceeding 50\%, showing that with examples, they can outperform larger models. 
% Consistent quality improvements were observed across all evaluated models and datasets. 

When we zoom into individual queries, we notice that \name improves the distribution of scores toward higher values. As shown in Figure \ref{fig:phi3_score}, without \name, small models frequently generated responses that received a score of -3 (significantly worse). With \name, we observe a marked reduction in responses scoring -3, with the overall distribution shifting rightward toward higher scores. The mean score improves substantially from -2.33 to -0.89, with nearly 50\% of queries performing at or above the level of large models.

We provide the side-by-side response quality comparison of generation scores among all three model families on five datasets in Figure \ref{fig:quality_improvements_raw_score}.

% \begin{figure}[t]
%     \centering
%     \includegraphics[width=0.8\linewidth]{Figures/wildchat_32b.png}
%     \caption{\name shifts traffic to higher throughput models under higher QPS.}
%     \label{fig:wildchat_qps_regime}
% \end{figure}

% \paragraph{Impact of Router}
% We evaluated our Request Router's  ability to dynamically allocate requests among multiple LLMs (Qwen 1.5B, 7B, 14B and 32B). The evaluation used English conversations from the Wildchat \cite{zhao2024wildchat} dataset, with response quality assessed by Gemini 2.5 Pro. The bandit policy, trained on a 100k sample subset as per the methodology described earlier, was tested on a distinct 10k sample set under varying QPS. Figure~\ref{fig:wildchat_qps_regime} indicates that at low QPS, the bandit primarily utilizes the 14B model which significantly outperforms the 32B model due to IC-Cache. As QPS increases, demonstrating its load-aware capability, the policy progressively shifts traffic towards models with higher throughput, predominantly routing to the 1.5B model under high load conditions to maintain system responsiveness. We note that the 14B model's throughput was constrained in this setup due to long prefill times from included examples; implementing KV caching, omitted in this evaluation, could improve its performance.